\documentclass{article}

\PassOptionsToPackage{numbers, compress}{natbib}

    \usepackage[preprint]{neurips_2025}

\usepackage[utf8]{inputenc} 
\usepackage[T1]{fontenc}    
\usepackage{hyperref}       
\usepackage{url}            
\usepackage{booktabs}       
\usepackage{amsfonts}       
\usepackage{nicefrac}       
\usepackage{microtype}      
\usepackage{xcolor}         
\usepackage{multirow}
\usepackage{xcolor}
\usepackage{colortbl}
\usepackage{algorithm}
\usepackage{algorithmicx}
\usepackage{algpseudocode}
\usepackage{amsmath}
\usepackage{amssymb}
\usepackage{graphics}
\usepackage{ProjLib}
\usepackage{subcaption}

\title{MathPhys-Guided Coarse-to-Fine Anomaly Synthesis with SQE-Driven Bi-Level Optimization for Anomaly Detection}

\author{
Long Qian$^{1,2}$~~~~
Bingke Zhu$^{1,3}$~~~~
Yingying Chen$^{1,3}$~~~~
Ming Tang$^{1,2}$~~~~
Jinqiao Wang$^{1,2,3}$\\
  $^{1}$~Foundation Model Research Center, Institute of Automation, \\ Chinese Academy of Sciences, Beijing, China \\ 
  $^{2}$~School of Artificial Intelligence, University of Chinese Academy of Sciences, Beijing, China\\
  $^{3}$~Objecteye Inc., Beijing, China\\
  {\tt\small  qianlong2024@ia.ac.cn} \\
  {\tt\small \{bingke.zhu,yingying.chen,tangm,jqwang\}@nlpr.ia.ac.cn}
}

\begin{document}

\maketitle

\begin{abstract}
    Currently, industrial anomaly detection suffers from two bottlenecks: (i) the rarity of real-world defect images and (ii) the opacity of sample quality when synthetic data are used. Existing synthetic strategies (\eg, cut-and-paste) overlook the underlying physical causes of defects, leading to inconsistent, low-fidelity anomalies that hamper model generalization to real-world complexities. In this paper, we introduce a novel and lightweight pipeline that generates synthetic anomalies through \textit{Math-Phys model} guidance, refines them via a \textit{Coarse-to-Fine} approach and employs a \textit{bi-level} optimization strategy with a Synthesis Quality Estimator (SQE). By combining physical modeling of the three most typical physics-driven defect mechanisms: Fracture Line (FL), Pitting Loss (PL), and Plastic Warpage (PW), our method produces realistic defect masks, which are subsequently enhanced in two phases. The first stage (\textit{npcF}) enforces a PDE-based consistency to achieve a globally coherent anomaly structure, while the second stage (\textit{npcF++}) further improves local fidelity. Additionally, we leverage SQE-driven weighting, ensuring that high-quality synthetic samples receive greater emphasis during training. To validate our method, we conduct experiments on three anomaly detection benchmarks: \textit{MVTec~AD}, \textit{VisA}, and \textit{BTAD}. Across these datasets, our method achieves state-of-the-art results in both image- and pixel-AUROC, confirming the effectiveness of our \textit{MaPhC2F} dataset and \textit{BiSQAD} method. All code will be released.
\end{abstract}

\section{Introduction}
\label{sec:intro}

Recent developments in deep learning have spurred substantial progress in anomaly detection, particularly through techniques that harness synthetic anomalies to compensate for the scarcity and variability of real-world defects. On one hand, self-supervised methods based on simple cut-and-paste perturbations~\cite{li2021cutpaste} or reverse distillation~\cite{Deng2022RD4AD, gu2024rethinking, qian2024friend} show that learning “normal” appearance can be highly effective without labeled anomalies. On the other hand, more advanced generation strategies have emerged in the past few years. Diffusion-based approaches~\cite{Wyatt2022AnoDDPM,Zhang2023DiffusionAD_ICCV,Mousakhan2023DDAD} and transformer-driven models enable controllable or prompt-guided synthesis of subtle, realistic defects. Yet these models still treat defects as generic texture edits, leaving the underlying physics untold.
\begin{figure}[t]
  \centering
  \hfill
  \begin{subfigure}[b]{0.49\linewidth}
    \centering
    \includegraphics[width=\linewidth]{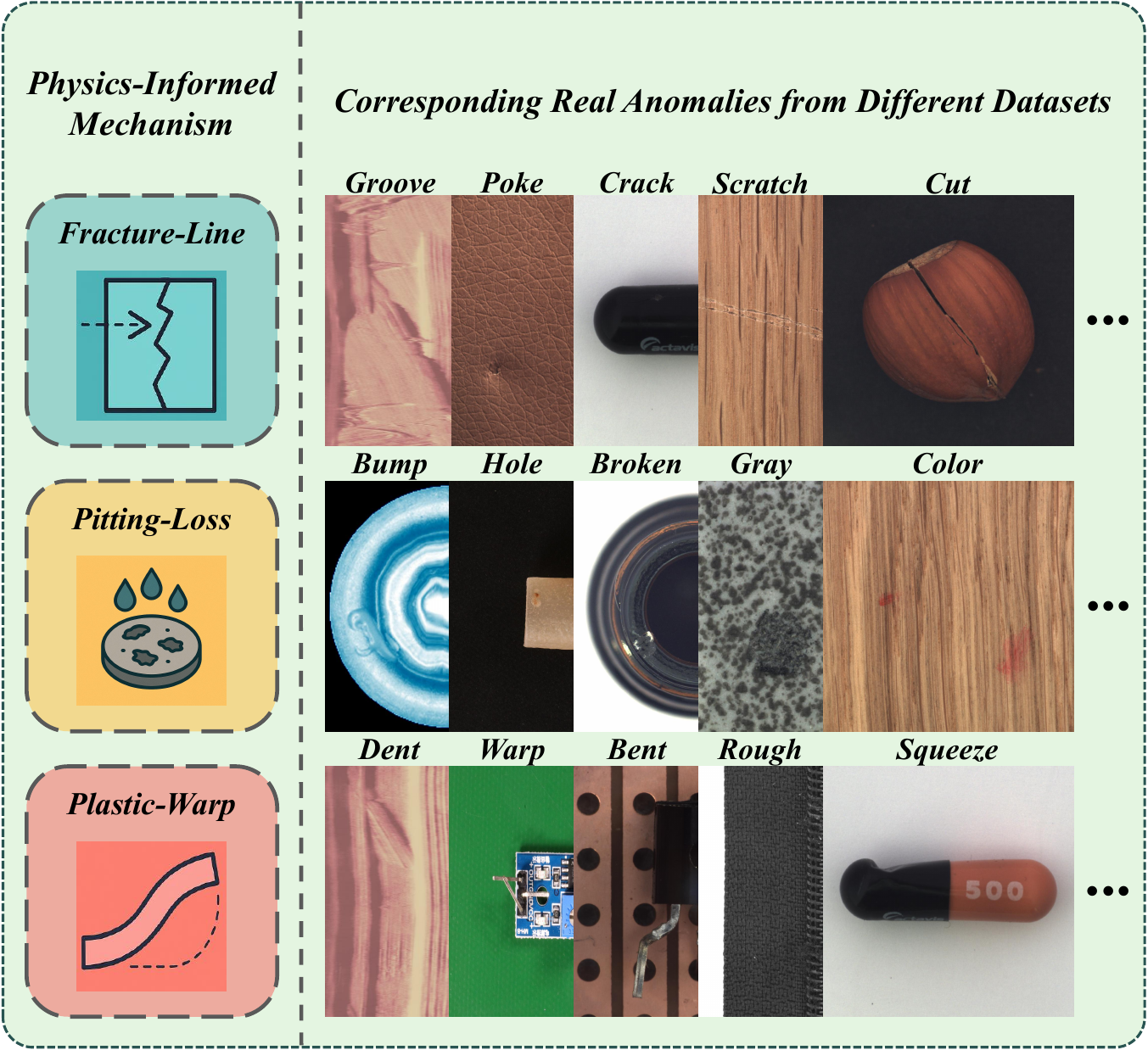}
    \caption{Mapping physics mechanisms to real anomalies.}
    \label{fig:synthetic_vs_real}
  \end{subfigure}
  \begin{subfigure}[b]{0.49\linewidth}
    \centering
    \includegraphics[width=\linewidth]{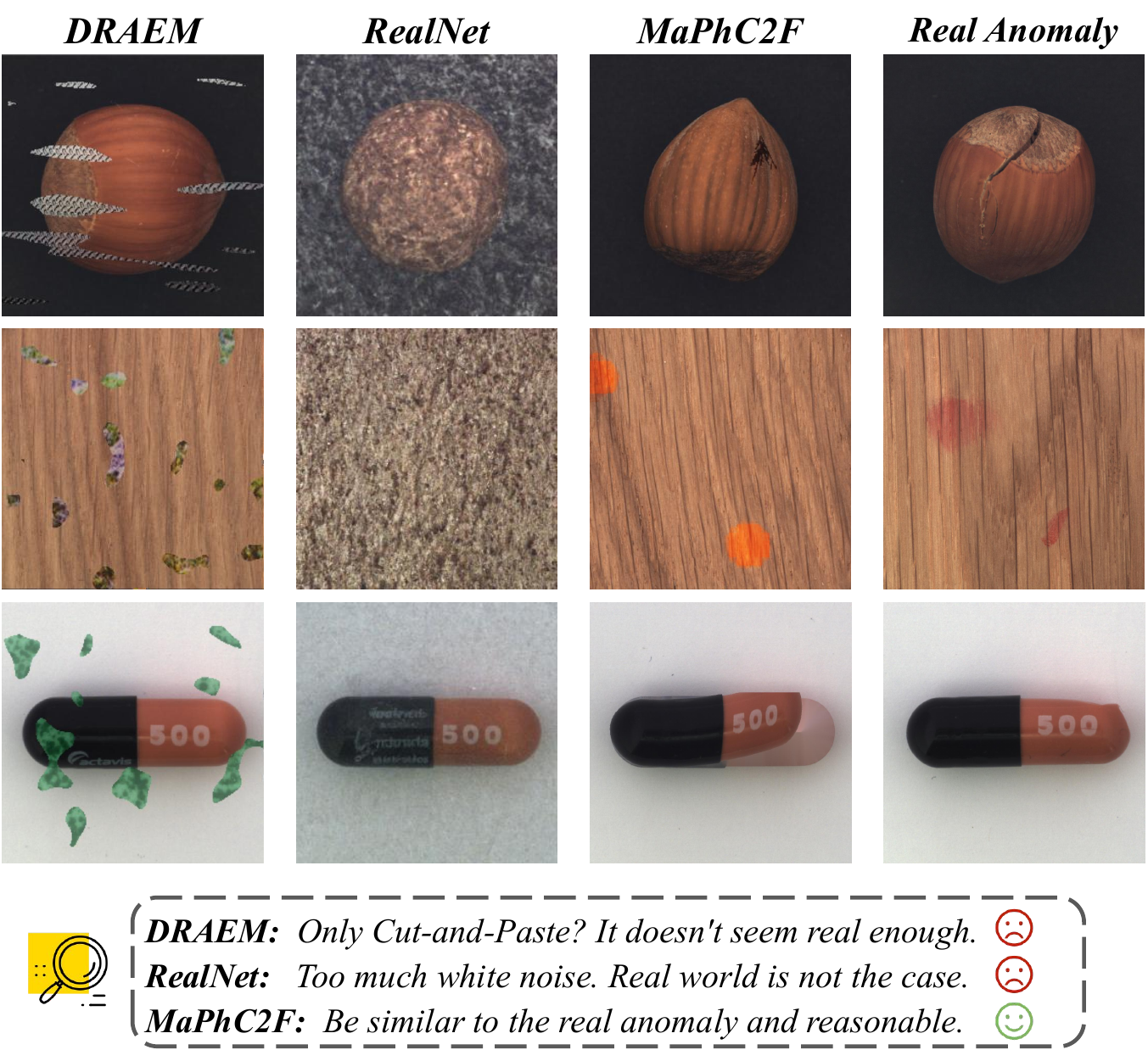}
    \caption{Comparison of anomaly synthesis methods.}
    \label{fig:anomaly_comparison}
  \end{subfigure}
  \begin{subfigure}[b]{0.98\linewidth}
    \centering
    \includegraphics[width=\linewidth]{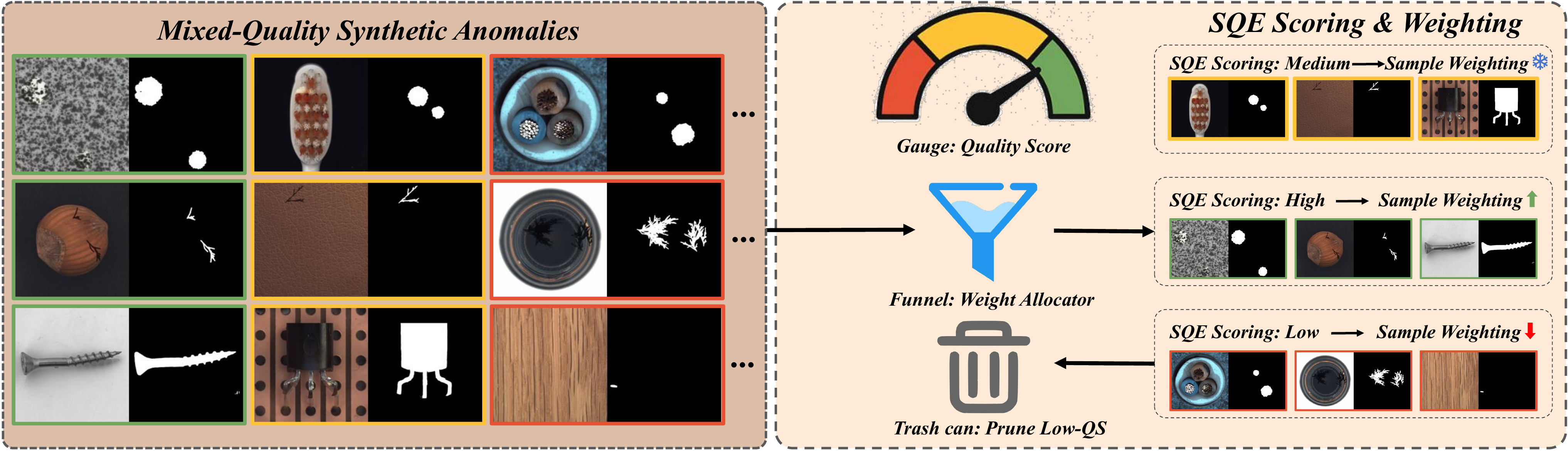}
    \caption{Motivation of \textit{SQE} automatic Scoring \& Weighting.}
    \label{fig:motivation}
  \end{subfigure}
    \caption{From Physics-Informed Anomaly Synthesis to SQE-driven Anomaly Detection: \textit{MaPhC2F} generates diverse anomalies, while \textit{SQE} up-weights the most reasonable ones.}
  \label{fig:triple_subfig}
\end{figure}
Despite the promise of synthetic anomaly data, two critical issues remain unsolved. \emph{First}, conventional synthetic strategies often fail to incorporate realistic physical phenomena due to lack of mathematical and physical model constraints, resulting in anomalies that lack fidelity to genuine defects. \emph{Second}, even if it becomes feasible to generate large volumes of anomalies, the results from previous methods can be of \emph{vastly varying quality}. Some synthesized defects deviate so strongly that they may degrade detection performance when used naively in training. A mechanism to \emph{evaluate} and \emph{selectively emphasize} high-quality synthesized data is thus sorely needed.


To address these challenges, we present a novel framework entitled \textit{\textbf{MathPhys-Guided Coarse-to-Fine}} anomaly synthesis method and \textit{\textbf{Bi-Loop SQE-Anomaly Detection}}. First, our \textbf{Ma}th\textbf{Ph}ys-Guided \textbf{C}oarse-to-\textbf{F}ine method, called \textit{\textbf{MaPhC2F}}, synthesizes anomalies by leveraging \emph{mathematically inspired models} that follow physical principles. Specifically, we generate defect masks based on math-phys models. These masks are then refined using a coarse-to-fine pipeline.

Physics lies at the core of how industrial surface anomalies arise. Defects rarely arise from random causes; each defect stems from a corresponding physical mechanism; they typically emerge from (1) tensile–stress fracture that opens linear fissures, (2) electro-chemical corrosion that excavates shallow pits, and (3) local plastic yield that warps geometry. Fig.~\ref{fig:synthetic_vs_real} shows correspondence between physics-informed mechanisms and real anomaly classes, demonstrating that the three most typical defect mechanisms effectively represent a wide range of real-world anomalies. Furthermore, our synthesis framework is highly extensible; additional anomaly categories can be introduced by defining new mask generators. In Fig.~\ref{fig:anomaly_comparison}, we illustrate how existing methods (\textit{DRAEM}~\cite{Zavrtanik2021DRAEM}, \textit{RealNet}~\cite{Zhang2024RealNet}) produce anomalies, while our \textit{MaPhC2F} generates results closer to real anomalies.


Furthermore, to address the quality differences among anomaly synthesis, we introduce a \textbf{Bi}-Loop \textbf{SQ}E-\textbf{A}nomaly \textbf{D}etection method, \textit{\textbf{BiSQAD}}, which unites a \emph{Synthesis Quality Estimator (SQE)} network with a \emph{bi-loop optimization} strategy. For the \textit{SQE}, we freeze a pretrained network to extract features for each synthetic sample and train only a lightweight final layer. After warmup, we record the per-sample losses from the model. These losses, after min--max normalization, serve as pseudo-labels for \textit{SQE}, indicating the quality of each synthetic example. 

Specifically, \textbf{BiSQAD} follows a meta-learning style bi-loop process. The \emph{inner loop} trains the anomaly detection model on both real and synthetic samples, recording each sample’s loss. Then, in the \emph{outer loop}, the model’s main parameters remain fixed while we update the SQE network (and optionally per-sample weights) based on these recorded losses. Consequently, high-SQE anomalies gain more training influence, while lower-fidelity samples are downweighted, thus guiding the system to focus on the most beneficial synthetic anomalies.

Across three widely recognized industrial benchmarks—\emph{MVTec~AD}, \emph{VisA}, and \emph{BTAD}—our framework achieves state-of-the-art performance in both image- and pixel-level AUROCs, indicating strong detection accuracy. Additionally, by curating and releasing a large-scale \textit{MaPhC2F Dataset} with 115{,}987 synthetic images, we further enrich the resources for anomaly detection research.

In essence, these contributions solve two critical problems with existing synthetic methods: (1) the inability to capture realistic anomalies and (2) the lack of a mechanism to evaluate and selectively focus on higher-quality generated anomalies. Our results demonstrate that integrating physically inspired anomaly masks and adaptive weighting boosts robustness against low-fidelity samples, consequently improving model performance. We summarize our main contributions as follows:

\begin{itemize}
    \item We propose a \emph{MathPhys-guided} method to anomaly synthesis, capturing diverse real-world defects by modelling their physical mechanisms and refining them via a lightweight coarse-to-fine process. Furthermore, we provide a \textbf{MaPhC2F Dataset} derived from this pipeline.
    \item We introduce \textbf{SQE-driven bi-loop} optimization (\textit{\textbf{BiSQAD}}), which learns to reweight synthetic data based on their fidelity and utility for anomaly detection, dynamically improving consistency to synthetic artifacts.
    \item Our framework integrates these components into an end-to-end pipeline. Extensive experiments on multiple benchmarks show \textbf{SOTA} performance, demonstrating significant improvements over prior synthetic-based anomaly detection methods.
\end{itemize}

\section{Related Work}
Recent years have focused on synthetic methods for anomaly detection, drawing upon advances in \emph{generative models}, \emph{self-supervised data augmentation}, and \emph{diffusion-based} approaches.

\paragraph{Self-Supervised Augmentation and Autoencoders.}
Another line of work employs \emph{autoencoders} and data augmentations to simulate anomalies for self-supervised tasks. \emph{CutPaste}~\cite{li2021cutpaste} and \emph{DRAEM}~\cite{Zavrtanik2021DRAEM} randomly splices image patches onto normal samples to produce pseudo-defects. Follow-up techniques integrate more advanced skills, as in \emph{NSA}~\cite{Schluter2022NSA}.

\paragraph{GAN.}
GAN have traditionally played a key role in synthesizing data for anomaly detection. Early works such as \emph{GANomaly}~\cite{Akcay2018GANomaly} and \emph{f-AnoGAN}~\cite{Schlegl2019fAnoGAN} introduced reconstruction-style pipelines, reconstructing normal data and flagging anomalies based on reconstruction discrepancies. Later, \emph{OCGAN}~\cite{Perera2019OCGAN} extended these ideas to detection with a latent-space regularization. GAN-based frameworks' training instability and limited coverage of the anomaly manifold remain open challenges~\cite{Pang2021Survey}.

\paragraph{Diffusion Models.}
Recently, \emph{diffusion} and \emph{score-based} models have emerged as powerful generators of synthetic anomalies. For example, \emph{AnoDDPM}~\cite{Wyatt2022AnoDDPM} introduced a denoising diffusion approach to produce subtle artifacts not captured by GANs. Later works leverage diffusion for both anomaly detection and synthesis, such as \emph{AnomalyDiffusion}~\cite{Hu2024AnomalyDiffusion} and \emph{TransFusion}~\cite{Fucka2024TransFusion}, which jointly learn to generate anomalies and discriminate them from normal data.


\paragraph{Hybrid Approaches and Advanced Architectures.}
Beyond pure synthesis or direct detection, new hybrid methods unify generation and anomaly detection. For instance, \emph{RealNet}~\cite{Zhang2024RealNet} proposed a feature selection backbone that can incorporate realistic synthetic defects to guide training. \emph{HardSynth}~\cite{Kim2024HardSynth}, \emph{AnoGen}~\cite{Gui2024AnoGen} attempts to generate hard samples that push detectors to improve. Meanwhile, refinement-based approaches (\eg., \ combining wavelet or PDE constraints~\cite{Fucka2024TransFusion, Du2022VOS}) produce higher fidelity anomalies that more closely mimic real anomalies.

Why existing synthetic pipelines fall short? Current methods treat defects as free-floating texture blobs, but these operations ignore load distribution, corrosion kinetics, or warp behaviour. In contrast to prior works, our pipeline distinguishes our method from purely generative methods, which typically lack \emph{physics principles} and mechanisms for identifying inferior synthetic data.

\section{Method}
\label{sec:method}
\subsection{Overview}
\begin{figure*}[t]
    \centering
    \includegraphics[width=\linewidth]{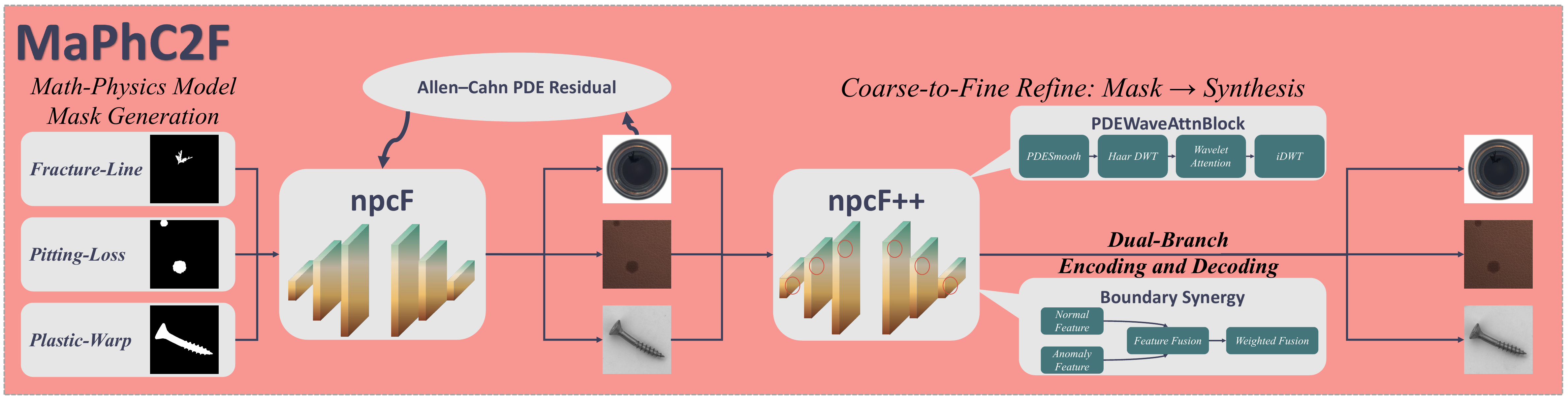}
    \caption{Track I – \textit{MaPhC2F}: MathPhys–guided coarse-to-fine anomaly synthesis.  Mask generators feed a coarse synthesiser (\textit{npcF}) and a fine refiner (\textit{npcF++}), producing photorealistic anomalies.}
    \label{fig:overall_maphc2f}
\end{figure*}
\begin{figure*}[t]
    \centering
    \includegraphics[width=\linewidth]{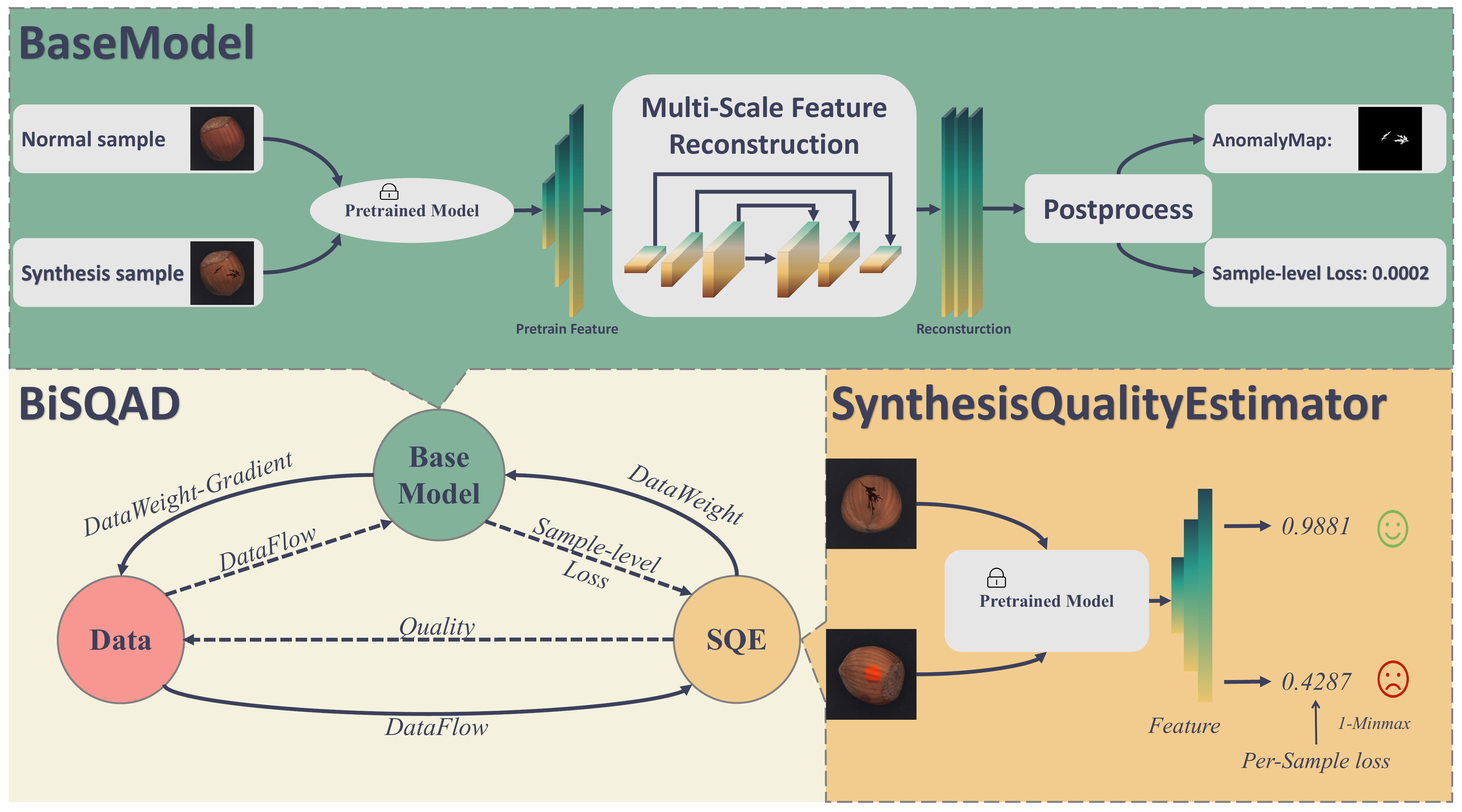}
    \caption{Track II – \textit{BiSQAD}: SQE-Driven Bi-Level Optimization for anomaly detection. Normal and synthetic images in Track I are fed into a base model while a learnable SQE assigns sample weights.}
    \label{fig:overall_pipeline}
\end{figure*}
We split the paper into two synchronized tracks (Fig.~\ref{fig:overall_maphc2f} and Fig.~\ref{fig:overall_pipeline}): Track I fabricates physically-sound anomalies, while Track II teaches the detector to trust only the sound ones. In Fig.~\ref{fig:overall_maphc2f} and Fig.~\ref{fig:overall_pipeline}, we present a concise overview of our entire framework, highlighting four key components: \emph{MaPhC2F}, \emph{BiSQAD}, \emph{BaseModel} and the \emph{SynthesisQualityEstimator (SQE)}. First, MaPhC2F produces math-physically informed anomalies via a coarse-to-fine refinement pipeline, which are then fed—together with normal samples—into the \textit{BaseModel} for anomaly detection. Here we use \textit{RealNet}~\cite{Zhang2024RealNet} as our \textit{BaseModel}. Next, \textit{BiSQAD} acts as a bi-loop optimization loop that balances each synthetic sample’s training influence based on the \textit{SQE}’s quality scores. Our method selectively highlights more useful anomalies while down-weighting weaker ones, boosting overall performance.

\subsection{\textit{MaPhC2F}: Anomaly Synthesis}
\subsubsection{Math-Physics Guided Anomaly Mask Synthesis}
\label{sec:math_phys_guided}
In this subsection, we describe how we generate three types of physically inspired anomaly masks---\emph{Fracture-Line}, \emph{Pitting-Loss}, and \emph{Plastic-Warp}. As a prerequisite, we use a segmentation model \textit{SAM}~\cite{kirillov2023segany} to extract the foreground region corresponding to the relevant item category, ensuring that our subsequent anomaly generation applies only within the object of interest.

\paragraph{Fracture-Line Mask Generation}
Guided by \emph{linear-elastic fracture mechanics}, we model Fracture-Line propagation on a 2D plane of size $H\times W$ by iteratively extending frontiers randomly. Let

\begin{equation}
\label{eq:crack_skeleton}
    \mathcal{S}(y,x) \;=\; 
    \begin{cases}
    1, & \text{if pixel $(y,x)$ lies on a Fracture-Line skeleton},\\
    0, & \text{otherwise}.
    \end{cases}
\end{equation}
For each frontier at position $(y,\,x)$ with movement vector $(\mathrm{d}y,\,\mathrm{d}x)$ and remaining steps $M>0$, we update
\begin{equation}
\label{eq:crack_step}
    y_{\mathrm{new}} \;=\; y + \mathrm{d}y,\quad
    x_{\mathrm{new}} \;=\; x + \mathrm{d}x,
\end{equation}
then mark $\mathcal{S}(y_{\mathrm{new}}, x_{\mathrm{new}})=1$ if it is within the foreground region. With probability $p_{\mathrm{branch}}$, a new branch is spawned by sampling a \emph{branch angle} $\Delta \theta$ in $\left[-\tfrac{\pi}{4},\,\tfrac{\pi}{4}\right]$ and rotating $(\mathrm{d}y,\,\mathrm{d}x)$ accordingly. Once a frontier’s step count is exhausted or a random stop threshold is met, it terminates. This yields a skeleton that may branch and meander, mimicking real-world Fracture-Line paths.

Next, to produce a binary Fracture-Line mask $\mathcal{M}_{\mathrm{FL}}(y,x)\in\{0,1\}$, we evaluate an exponential threshold around each skeleton pixel based on its distance transform. Let
\begin{equation}
\label{eq:dist_transform}
  \mathrm{dist}(y,x) \;=\; \text{distance\_transform}\!\bigl(1 - \mathcal{S}(y,x)\bigr),
\end{equation}
where large $\mathrm{dist}(y,x)$ indicates a point far from the skeleton. We define a local thickness function:
\begin{equation}
\label{eq:thickness}
  w_{t}(y,x) \;=\; w_0 \exp\!\bigl(-\alpha\,\mathrm{dist}(y,x)\bigr)\;+\;\varepsilon,
\end{equation}
where $w_0>0$, $\alpha>0$, and $\varepsilon>0$ are user-chosen parameters. In parallel, a spatially varying Perlin noise $\nu(y,x)$ is sampled:
\begin{equation}
\label{eq:perlin}
  \nu(y,x) \;=\; \mathrm{PN}\!\Bigl(\tfrac{y}{H}, \tfrac{x}{W}\Bigr)\,\cdot\,\mathrm{scale},
\end{equation}
with \(\mathrm{PN}\) denoting PerlinNoise(\(\cdot\)) and $\mathrm{scale}>0$. If
\begin{equation}
\label{eq:mask_condition}
  \mathrm{dist}(y,x)\;+\;\nu(y,x)\;<\;w_{t}(y,x),
\end{equation}
then $\mathcal{M}_{\mathrm{FL}}(y,x)=1$; otherwise 0. This procedure ensures that regions near the skeleton remain inside the Fracture-Line, whereas distant regions are excluded.

\paragraph{Pitting-Loss Mask Generation}
We simulate blocky or aggregated Pitting-Loss patterns by randomly placing polygons within the foreground mask and then growing them via boundary expansions. Let $\mathcal{M}_{\mathrm{PL}}(y,x)$ be the Pitting-Loss mask. We begin by sampling $k$ random polygons, each clipped to the foreground. These polygons may overlap (with probability $p_{\mathrm{overlap}}$) or be unioned otherwise. In each polygon, we fill internal pixels to 1. We then perform a local BFS-like boundary growth for $N_{\mathrm{growth}}$ iterations, turning boundary pixels into mask pixels with probability $p_{\mathrm{grow}}$. After morphological closing, an optional Perlin noise step randomly erodes boundary pixels:
\begin{equation}
\label{eq:perlin_corros}
  \textstyle
  \mathcal{M}_{\mathrm{PL}}(y,x)\;=\;
  \begin{cases}
    0, & \mbox{if }\nu\bigl(\tfrac{y}{H}, \tfrac{x}{W}\bigr)>\rho,\\
    \mathcal{M}_{\mathrm{PL}}(y,x), & \mbox{otherwise},
  \end{cases}
\end{equation}
where \(\rho\) is a threshold controlling edge breakage.

\paragraph{Plastic-Warp Mask Generation via Thin-Plate Spline}

To simulate local warping or bending, we adopt a thin-plate spline (TPS) approach. Let $\mathcal{M}_{\mathrm{PW}}(y,x)$ represent the region to be deformed. After extracting a bounding box on the foreground, we sample \emph{control points} \(\{(r_i,\,c_i)\}_{i=1}^m\) along $\mathcal{M}_{\mathrm{PW}}$ and displace them by random offsets. An RBF-based TPS function~\cite{https://doi.org/10.1029/JB076i008p01905} is then computed to map each pixel $(y,x)$ to a new location $(y',\,x')$. Formally,
\begin{equation}
\label{eq:rbf_tps}
  (x',y') \;=\; (x,y)\;-\;\bigl[\mathrm{Rbf}_x(x,y), \;\mathrm{Rbf}_y(x,y)\bigr],
\end{equation}
where \(\mathrm{Rbf}_x\) and \(\mathrm{Rbf}_y\) are trained via radial basis functions to interpolate the control points. We apply this mapping to both the image and the mask, optionally inpainting the removed object area before overlay. This yields realistic local distortions akin to dents or folds.

Overall, these three math-physics guided processes produce diverse and physically plausible anomaly masks. In the following subsections, these masks serve as inputs to our coarse-to-fine refinement pipeline, ultimately generating high-fidelity anomalies for subsequent model training.

\subsubsection{\textit{npcF} (Coarse Refinement)}
\label{sec:npcf}

The coarse refinement, \textbf{\textit{npcF}}, applies a WideResNet autoencoder with \emph{Allen--Cahn PDE} constraints~\cite{allen1979microscopic, xiang2022self}, and other auxiliary terms to refine anomaly masks into a coherent form, smooth out unnatural boundaries, and avoid extreme color artifacts before proceeding to the fine-grained stage.

\paragraph{Allen--Cahn PDE Residual.}
We treat \(\mathbf{u}\in\mathbb{R}^{B\times 3\times H\times W}\) as a \emph{phase variable} that should stabilise into two states—background and defect. Accordingly we penalise the Allen--Cahn residual,
\begin{equation}
\label{eq:allen_cahn}
  \mathrm{Res}_{\mathrm{AC}}(\mathbf{u}) \;=\; \varepsilon^{2}\,\nabla^{2}\mathbf{u}\;-\;\Bigl(\mathbf{u}^{3}\;-\;\mathbf{u}\Bigr),
\end{equation}
where \(\nabla^{2}\) denotes the discrete 2D Laplacian operator, and \(\varepsilon>0\) is a parameter that controls the diffusion strength. Then we apply the Allen--Cahn loss restricted to anomaly regions \(\mathbf{M}_{\mathrm{anom}}\):
\begin{equation}
\label{eq:pde_loss}
  \ell_{\mathrm{pde}} \;=\; \bigl\|\mathrm{Res}_{\mathrm{AC}}(\mathbf{u})\;\odot\;\mathbf{M}_{\mathrm{anom}}\bigr\|^{2},
\end{equation}
where $\odot$ denotes elementwise multiplication. This drives pixel values toward the binary wells \(\{-1,+1\}\) while the \(\varepsilon^{2}\nabla^{2}\) term suppresses color overshoot, yielding sharper, materially consistent defect boundaries across different base textures.

\paragraph{Reconstruction \& Regularization}
\label{sec:npcf_short}
In \textit{npcf}, we apply separate reconstruction terms for normal and anomalous regions, combined with auxiliary regularizers. These ensure that reconstructed anomalies remain visually consistent. \emph{For the precise definitions, please see the ~\cref{sec:npcf_add} and ~\ref{sec:subsec_npcf}.}

\subsubsection{\textit{npcF++} (Fine-Grained Refinement)}
\label{sec:npcfpp}

After obtaining a coarse anomaly approximation via \textit{npcf}, we further refine local details, boundary transitions, and high-frequency texture in a process we call \textbf{\textit{npcF++}}.

\paragraph{Wavelet-PDE Attentive Blocks, Boundary Synergy, and Dual-Branch Design}
\label{sec:short_wpde_synergy}
We refine feature maps via a combination of \emph{PDEWaveAttnBlock} and \emph{RegionSynergyBlock}. These modules are organized in dual-branch encoders for normal and anomaly inputs, followed by a shared bottleneck and decoding path. \emph{For precise definitions, including how wavelet decomposition and boundary cross-attention are integrated across multiple scales, see the ~\cref{sec:npcfpp_supp} and ~\ref{sec:subsec_npcfpp}.}

\paragraph{Loss Functions.}
Let \(\mathbf{u}\in\mathbb{R}^{B\times 3\times H\times W}\) represent a batch of reconstructed images output by the autoencoder. We rely on a \emph{region loss} and a \emph{wavelet high-frequency} penalty:

\begin{equation}
\label{eq:region_loss}
\resizebox{.9\linewidth}{!}{$
  \ell_{\mathrm{region}} = 
  \|(\mathbf{u} \odot \mathbf{M}_{\mathrm{norm}})-(\mathbf{x}_{\mathrm{orig}} \odot \mathbf{M}_{\mathrm{norm}})\|_{1} + \beta \Bigl( \|(\mathbf{u} \odot \mathbf{M}_{\mathrm{anom}}) - (\mathbf{x}_{\mathrm{b1}} \odot \mathbf{M}_{\mathrm{anom}})\|_{1} + \delta \|\mathbf{u} - \mathbf{x}_{\mathrm{orig}}\|_{2}^{2} \Bigr)
$}
\end{equation}

where $\mathbf{M}_{\mathrm{norm}}=1-\mathbf{M}_{\mathrm{anom}}$, $\mathbf{x}_{\mathrm{b1}}$ is the coarse anomaly image, and \(\beta,\delta\) are weighting factors. Additionally, we penalize high-frequency edges in anomaly regions via:
\begin{equation}
\label{eq:wave_hf}
  \ell_{\mathrm{waveHF}}
  \;=\;
  \textstyle
  \bigl\|
    \mathrm{convHF}\!\bigl(\mathbf{u}\bigr)
    \;\odot\;
    \mathbf{M}_{\mathrm{anom}}
  \bigr\|_{1},
\end{equation}
where \(\mathrm{convHF}(\cdot)\) is a $3\times 3$ \textit{Laplacian} kernel~\cite{marr1980theory}. The total loss is then $\ell_{\mathrm{region}} + \ell_{\mathrm{waveHF}}$.

\subsection{\textit{BiSQAD}: SQE-Driven Bi-Level Optimization for Anomaly Detection}
\label{sec:biSQAD}

With refined anomalies, we tackle the issue of varying synthetic defect quality, as some samples closely match real defects while others could hinder detection. Then we propose bi-loop optimization, \textbf{BiSQAD}, that leverages a \emph{Synthesis Quality Estimator} to dynamically weight samples' contribution.

\subsubsection{Synthesis Quality Estimator}
\label{sec:sqe}

We define a scoring function $q_{i}\in[0,1]$ for each synthetic sample $i$. Let $\mathbf{x}_{\mathrm{sqe}}$ be the normalized image patch fed to a frozen backbone WideResNet-50 followed by a learnable final layer:
\begin{equation}
    \label{eq:SQE}
    q_{i}
    \;=\;
    \sigma\Bigl(
      \mathrm{fc}\Bigl(
        \mathrm{FeatExtract}\bigl(\mathbf{x}_{\mathrm{sqe}}\bigr)
      \Bigr)
    \Bigr),
\end{equation}
where $\sigma(\cdot)$ is a sigmoid function. To train SQE \emph{without} manual labels, we observe basemodel’s per-sample losses $\ell_i$ during training. The target quality for sample $i$ is
\begin{equation}
    \label{eq:SQE_target}
    y_{i}
    \;=\;
    1.0 \;-\;\frac{\ell_i - \min(\ell)}{\max(\ell)-\min(\ell)+\varepsilon},
\end{equation}
where $\varepsilon$ is a small constant preventing division by zero. We then regress $q_i$ toward $y_i$ using an $\ell_2$ loss. As a result, synthetic anomalies that produce higher loss are assigned lower quality $q_i$, and vice versa.

\subsubsection{Bi-Loop Optimization}
\label{sec:bilevel_optim}

\paragraph{Inner Loop: Loss-weight Training.}
During training, each sample $i$ is assigned a dynamic weight:
\begin{equation}
\label{eq:bilevel_weight}
  w_i \;=\; \lambda_{\mathrm{sqe}} \, q_i 
  \;+\;
  \lambda_{\mathrm{bi}}\,d_i,
\end{equation}
where $d_i$ is an optional learnable data weight and $\lambda_{\mathrm{sqe}},\lambda_{\mathrm{bi}}$ are hyperparameters. We scale the main detection loss by $w_i$, generating an updated model parameter set $\theta'$. 

\paragraph{Outer Loop: Data-weight's Learning.}
At the end of each epoch, we refine a SQE layer as shown in ~\cref{sec:sqe} and the sample-level data weights (\(d_i\)) using a validation-based objective. This optional, strictly second-order MAML step temporarily freezes the main model $\theta'$ and backpropagates through the inner-loop graph to adjust \(\{d_i\}\). \emph{Further details and full equations appear in the ~\cref{sec:outer_loop_supp}.}

In summary, \emph{BiSQAD} ties together an online SQE with a second-order meta-learning loop. This synergy ensures that training continuously promotes anomalies beneficial to the detector and curbs those that risk hurting final performance.

\section{Experiment}
\label{sec:exp}
\subsection{Experiment Setup}

\paragraph{Datasets.}In this work, we evaluate our approach on three popular anomaly detection datasets: \textit{MVTec~AD}~\cite{bergmann2019mvtec}, \textit{VisA}~\cite{zou2022spot} and \textit{BTAD}~\cite{9576231}. \textit{MVTec~AD}  contains 5,354 images across 15 categories, \textit{VisA} comprises 10,821 images spanning 12 categories and \textit{BTAD} contains 2540 images of 3 categories.
\paragraph{Implementation details.}For most stages of our pipeline, we set the batch size to 32 and train for about 200 epochs on the coarse refinement(\textit{npcF}) and 50 epochs on the fine refinement(\textit{npcF++}). During training, images are resized to $512\times512$ resolution. More Implementation details, and parameter selection criteria are shown in ~\cref{sec:psc} 


\paragraph{Metrics.}We quantify anomaly detection performance using the image- and pixel-level AUROC.

\subsection{Comparison with SOTAs}
\label{sec:comparison_sota}
\begin{figure}[h]
    \centering
    \includegraphics[width=1\linewidth]{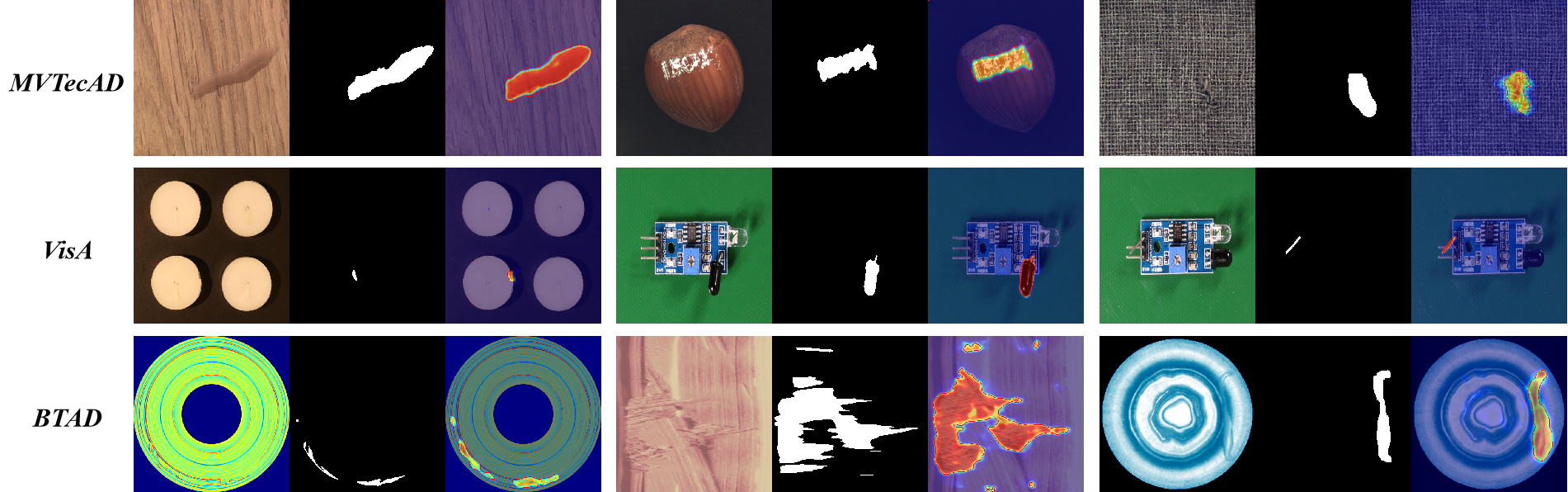}
    \caption{Visualization examples of anomaly detection results on \emph{MVTec~AD}, \emph{VisA}, and \emph{BTAD}.}
    \label{fig:vis}
\end{figure}
\begin{table*}[h]
\centering
\resizebox{\textwidth}{!}{
\begin{tabular}{c|c|ccccccc}
\toprule
\multicolumn{2}{c|}{Category} & \textit{DSR}~\cite{10.1007/978-3-031-19821-2_31} & \textit{CutPaste}~\cite{li2021cutpaste}
& \textit{DRAEM}~\cite{Zavrtanik2021DRAEM} & \textit{AnomalyDiffusion}~\cite{Hu2024AnomalyDiffusion} & \textit{RD++}~\cite{Tien_2023_CVPR}  & \textit{RealNet}~\cite{Zhang2024RealNet}
 & \textit{BiSQAD}
\\
\midrule
\multirow{10}{*}{\rotatebox{90}{Object}} 
& \textit{Bottle}     & 99.6/98.8 & \textbf{100}/99.1
& 99.2/97.8 & 99.8/99.4 & \textbf{100}/98.8  & \textbf{100}/99.3
 & \textbf{100}/\textbf{99.6}
\\
& \textit{Cable}      & 95.3/97.7 & 96.4/96.2
& 91.8/94.7 & \textbf{100}/99.2 & 99.3/98.4  &99.2/98.1
 &99.6/\textbf{99.6}
\\
& \textit{Capsule}    & 98.3/91.0 & 98.5/99.1
& 98.5/94.3 & \textbf{99.7}/98.8 & 99.0/98.8  &99.6/99.3
 &99.4/\textbf{99.6}
\\
& \textit{Hazelnut}   & 97.7/99.1 & \textbf{100}/99.0
& \textbf{100}/99.7 & 99.8/99.8 & \textbf{100}/99.2  &\textbf{100}/99.8&\textbf{100}/\textbf{99.9}
\\
& \textit{Metal Nut}   & 99.1/94.1 & 99.9/98.0
& 98.7/99.5 & \textbf{100}/99.8& \textbf{100}/98.1  &99.7/98.6
 &99.7/99.4
\\
& \textit{Pill}       & 98.9/94.2 & 97.2/99.0
& 98.9/97.6 & 98.0/\textbf{99.8} & 98.4/98.3  &\textbf{99.1}/99.0
 &98.2/\textbf{99.8}
\\
& \textit{Screw}      & 95.9/98.1 & 92.7/98.5
& 93.9/97.6 & 96.8/97.0 & \textbf{98.9}/99.7  &98.8/99.5
 &98.1/\textbf{99.8}
\\
& \textit{Toothbrush} & \textbf{100}/\textbf{99.5} & 99.2/98.9
& \textbf{100}/98.1 & \textbf{100}/99.2 & \textbf{100}/99.1  &99.4/98.7
 &99.4/98.7
\\
& \textit{Transistor} & 96.3/80.3 & 99.4/96.3
& 93.1/90.9 & \textbf{100}/\textbf{99.3} & 98.5/94.3  &\textbf{100}/98.0 & 99.4/98.1
\\
& \textit{Zipper}     & 98.5/98.4 & 99.6/98.0
& \textbf{100}/98.8 & 99.9/99.4 & 98.6/98.8  &99.8/99.2
 &\textbf{100}/\textbf{99.7}\\
\midrule
\multirow{5}{*}{\rotatebox{90}{Texture}} 
& \textit{Carpet}     & 99.6/96.0 & 99.2/98.4
& 97.0/95.5 & 96.7/98.6 & \textbf{100}/99.2  &99.8/99.2
 &99.7/\textbf{99.9}
\\
& \textit{Grid}       & \textbf{100}/99.6 & \textbf{100}/99.2
& 99.9/99.7 & 98.4/98.3 & \textbf{100}/99.3  &\textbf{100}/99.5&\textbf{100}/\textbf{99.8}\\
& \textit{Leather}    & 99.3/99.5 & \textbf{100}/99.4
& \textbf{100}/98.6 & \textbf{100}/\textbf{99.8} & \textbf{100}/99.5  &\textbf{100}/\textbf{99.8}&\textbf{100}/\textbf{99.8}
\\
& \textit{Tile}       & \textbf{100}/98.6 & 99.9/97.6
& 99.6/99.2 & \textbf{100}/99.2 & 99.7/96.6  &99.9/99.4
 &99.9/\textbf{99.7}
\\
& \textit{Wood}       & 94.7/91.5 & 99.0/95.0
& 99.1/96.4 & 98.4/\textbf{98.9} & \textbf{99.3}/95.8  &99.2/98.2
 &98.3/98.8
\\
\midrule
\rowcolor[gray]{0.9}
\multicolumn{2}{c|}{\textbf{\textit{Avg.}}}    & 98.2/95.8 & 98.7/98.1
& 99.2/99.1 & 98.5/97.7 & 99.4/98.3  &\textbf{99.6}/99.0
 &99.5/\textbf{99.5}
\\
\bottomrule
\end{tabular}
}
\caption{Performance comparison across different SOTA methods on \textit{MVTec~AD} dataset. \textbf{Bold text} indicates the best performance among all method. The values in the form of \textit{xx/xx} represent \textit{image-level AUROC / pixel-level AUROC}.}
\label{tab:mvtec}
\end{table*}

\begin{table*}[t]
\centering
\resizebox{\textwidth}{!}{
\begin{tabular}{c|ccccccc}
\toprule
Category & \textit{DSR}~\cite{cohen2020sub}& \textit{SimpleNet}~\cite{liu2023simplenetsimplenetworkimage}& \textit{DRAEM}~\cite{Zavrtanik2021DRAEM}& \textit{PatchCore}~\cite{Roth2022PatchCore} &\textit{RD++}~\cite{Tien_2023_CVPR}& \textit{RealNet}~\cite{Zhang2024RealNet}& \textit{BiSQAD}
\\
\midrule
\textit{Candle}& 86.4/79.7& 92.3/97.7& 91.8/96.6& \textbf{98.6}/99.5 &96.4/98.6& 96.1/99.1& 96.5/\textbf{99.8}
\\
\textit{Capsules} & 93.4/74.5& 76.2/94.6& 74.7/98.5& 81.6/\textbf{99.5} &92.1/99.4& 93.2/98.7&\textbf{95.4}/98.6
\\
\textit{Cashew} & 85.2/61.5& 94.1/99.4& 95.1/83.5& 97.3/98.9 &\textbf{97.8}/95.8& \textbf{97.8}/98.3&97.1/\textbf{99.9}
\\
\textit{Chewinggum} & 97.2/58.2& 97.1/97.0& 94.8/96.8& 99.1/99.1 &96.4/99.0& \textbf{99.9}/\textbf{99.8}&99.5/99.7
\\
\textit{Fryum} & 93.0/65.5& 88.0/93.5& \textbf{97.4}/87.2& 96.2/93.8 &95.8/94.3& 97.1/\textbf{96.2}&96.2/95.0
\\
\textit{Macaroni1} & 91.7/57.7& 84.7/95.4& 97.2/\textbf{99.9}& 97.5/99.8 &94.0/99.7& \textbf{99.8}/\textbf{99.9}&96.5/\textbf{99.9}
\\
\textit{Macaroni2} & 79.0/52.2& 75.0/83.8& 85.0/99.2& 78.1/99.1 &88.0/87.7& 95.2/\textbf{99.6}&\textbf{96.6}/98.4
\\

\textit{PCB1} & 89.1/61.3& 93.4/99.1& 47.6/88.7& \textbf{98.5}/\textbf{99.9} &97.0/75.0& \textbf{98.5}/99.7&98.1/\textbf{99.9}
\\
\textit{PCB2} & 96.4/84.9& 90.0/94.8& 89.8/91.3& 97.3/99.0 &97.2/64.8& \textbf{97.6}/98.0&96.5/\textbf{99.7}
\\
\textit{PCB3} & 97.0/79.5& 91.3/98.2& 92.0/98.0& 97.9/99.2 &96.8/95.5& \textbf{99.1}/98.8&98.8/\textbf{99.9}
\\
\textit{PCB4} & 98.5/62.1& 99.1/94.5& 98.6/96.8& 99.6/98.6 &\textbf{99.8}/92
.8& 99.7/98.6&99.6/\textbf{99.8}
\\
\textit{Pipe fryum} & 94.3/80.5& 89.0/95.3& \textbf{100.0}/85.8& 99.8/99.1 &99.6/92.0& 99.9/99.2&98.3/\textbf{99.8}
\\
\midrule

\rowcolor[gray]{0.9}
\textit{\textbf{Avg.}}    & 91.8/68.1& 89.2/95.3& 88.7/93.5& 95.1/98.8 &95.9/90.1& \textbf{97.8}/98.8&97.4/\textbf{99.2}
\\
\bottomrule
\end{tabular}
}
\caption{Performance comparison across different SOTA methods on \textit{VisA} dataset.}
\label{tab:visa}
\end{table*}

\begin{table*}[h]
\centering
\resizebox{\linewidth}{!}{
\begin{tabular}{c|cccccc}
\toprule
\textit{Category}& \textit{SimpleNet}~\cite{liu2023simplenetsimplenetworkimage}& \textit{SPADE}~\cite{cohen2020sub}& \textit{RD}~\cite{Deng2022RD4AD} &\textit{RD++}~\cite{Tien_2023_CVPR}& \textit{RealNet}~\cite{Zhang2024RealNet} & \textit{BiSQAD}
\\
\midrule
\textit{01} & 96.4/90.3& 91.4/97.3& 97.9/99.3 &96.8/96.2& \textbf{100}/\textbf{98.2}& 99.8/98.0
\\
\textit{02} & 75.2/48.9& 71.4/94.4& 86.0/97.7 &\textbf{90.1}/96.4& 88.6/96.3&89.1/\textbf{96.6}
\\
\textit{03} & 99.3/97.2& 99.9/99.1& 99.7/94.2 &\textbf{100}/\textbf{99.7}& 99.6/99.4&99.8/99.6
\\
\midrule
\rowcolor[gray]{0.9}
\textbf{\textit{Avg.}}& 90.3/78.8& 87.6/96.9 & 94.5/97.1  &95.6/97.4& 96.1/97.9 & \textbf{96.3}/\textbf{98.1}
\\
\bottomrule
\end{tabular}
}
\caption{Performance comparison across different SOTA methods on \textit{BTAD} dataset.}
\label{tab:btad}
\end{table*}

We benchmark our method on three prominent industrial anomaly detection datasets: \emph{MVTec~AD}, \emph{VisA}, and \emph{BTAD}. As shown in Table~\ref{tab:mvtec}, \textit{BiSQAD} achieves an overall \emph{99.5\%/99.5\%}on \textit{MVTec~AD}, outperforming other leading approaches in both \emph{object} and \emph{texture} categories. In Table~\ref{tab:visa}, our method obtains an average \emph{97.4\%/99.2\%} on \textit{VisA} and Table~\ref{tab:btad} reports results on \textit{BTAD}, where \textit{BiSQAD} again surpasses recent baselines, achieving \emph{96.3\%/98.1\%}. The result empirically supports the generalizability claims made in the Fig.~\ref{fig:synthetic_vs_real} and Table~\ref{tab:synthetic_vs_real}, demonstrating that our synthesized anomalies effectively represent diverse real-world defect scenarios across multiple datasets.

Visualization examples of anomaly detection on \textit{MVTec~AD}, \textit{VisA}, and \textit{BTAD} are presented in Fig.~\ref{fig:vis}. These results highlight the method’s fine-grained localization capabilities, identifying subtle boundary details and diverse defect patterns. More qualitative results in the ~\cref{sec:qualitative_results} confirm the robustness and SOTA performance of our method across various anomaly types.

Our method also demonstrates a clear advantage in terms of computational complexity and efficiency compared to existing methods, which we quantitatively analyze in detail in ~\cref{sec:complexity}.

\subsection{Ablation Study}

\begin{table}[h]
    \centering
    \begin{subtable}{0.49\textwidth}
        \centering
        \resizebox{\linewidth}{!}{
            \begin{tabular}{lccc}
                \toprule
                \textbf{Method Variant} & \textbf{Image-AUROC} & \textbf{Pixel-AUROC} & \textbf{Comments}\\
                \midrule
                (1) \textit{npcF} only   & 94.9 & 95.4 & - \\
                (2) +PDEWave             & 97.3(\textcolor{red}{+2.4}) & 97.2(\textcolor{red}{+1.8}) & partial fine stage\\
                \rowcolor[gray]{0.9}
                (3) Full \textit{npcF++} & \textbf{99.5}(\textcolor{red}{+2.2}) & \textbf{99.5}(\textcolor{red}{+2.3}) & PDEwave + synergy \\
                \bottomrule
            \end{tabular}
        }
        \caption{Ablation of \textit{npcF++}.}
        \label{tab:ablation_npcfpp}
    \end{subtable}
    \hfill
    \begin{subtable}{0.49\textwidth}
        \centering
        \resizebox{\linewidth}{!}{
            \begin{tabular}{lccc}
                \toprule
                \textbf{\textit{SQE} Variant} & \textbf{Image-AUROC} & \textbf{Pixel-AUROC} & \textbf{Comments}\\
                \midrule
                No \textit{SQE} & 95.6 & 96.4 & Default Weight\\
                \rowcolor[gray]{0.9}
                +\textit{SQE} & 99.5(\textcolor{red}{+3.9}) & 99.5(\textcolor{red}{+3.1}) & Score Weight\\
                \bottomrule
            \end{tabular}
        }
        \caption{Ablation of \textit{SQE}.}
        \label{tab:ablation_sqe}
    \end{subtable}
    \caption{Ablation results on \textit{MVTec AD} dataset.}
    \label{tab:ablation_combined}
\end{table}

        

        

We conduct ablation experiments on two core elements of our pipeline: \textit{npcF++} (within \textit{MaPhC2F}) and the SQE-driven reweighting in \textit{BiSQAD}. Table~\ref{tab:ablation_npcfpp} illustrates the benefit of refining our coarse stage (\textit{npcF}) with PDEWave-based operations and synergy blocks. Transitioning from \textit{npcF} alone to a partial refinement yields \emph{+2.4\%/+1.8\%} in image- and pixel-level AUROC. By enabling the complete \textit{npcF++} pipeline, we observe an additional \emph{+2.2\%/+2.3\%} gain, respectively, indicating that synergy blocks deliver fine-grained improvements in anomaly realism. Next, Table~\ref{tab:ablation_sqe} examines how our \textit{SQE} influences performance within \textit{BiSQAD}. Disabling \textit{SQE} reduces overall accuracy, whereas applying SQE-based score weighting improves image- and pixel-level AUROC by \emph{+3.9\%/+3.1\%}.

Altogether, these results confirm that \textit{npcF++} notably elevates the realism of generated anomalies by capturing subtle textures and boundary details, while the \textit{SQE}-driven weighting mechanism effectively mitigates the impact of low-fidelity synthetic samples.

\subsection{MaPhC2F Dataset}
\label{sec:maphc2f_dataset}
\begin{figure}[H]
    \centering
    \includegraphics[width=0.9\linewidth]{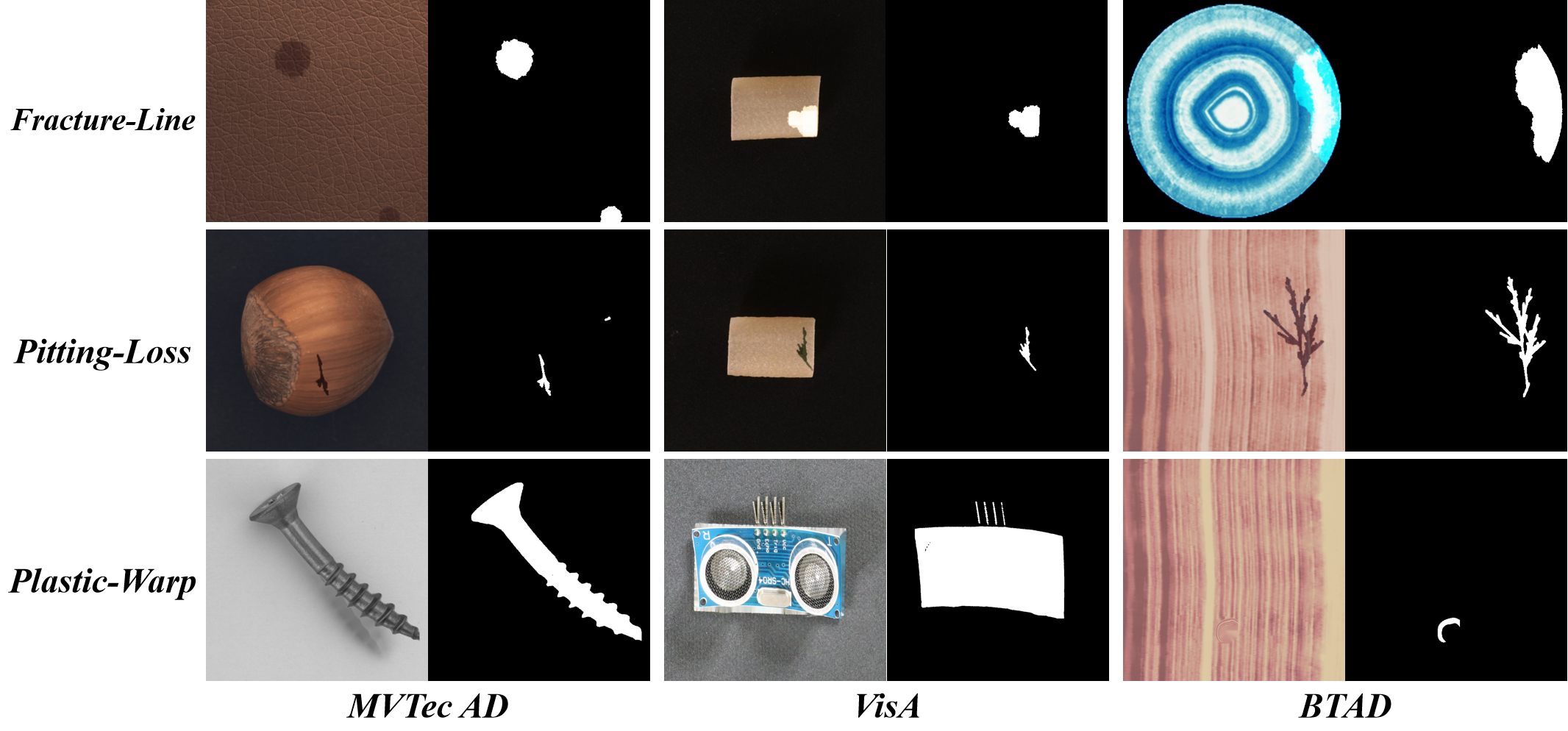}
    \caption{Visualization of three math-physics anomaly mechanisms in our \textbf{MaPhC2F Dataset}.}
    \label{fig:maphc2f_dataset}
\end{figure}

We introduce the \textbf{\textit{MaPhC2F Dataset}}, a large-scale synthetic anomaly collection generated via our math-physics guided pipeline. It spans \emph{115,987} images across \emph{30} object categories, each refined through the coarse-to-fine approach described in \cref{sec:npcf}--\cref{sec:npcfpp}. As depicted in Fig.~\ref{fig:maphc2f_dataset}, the dataset includes three representative anomaly mechanisms—\textit{Pitting-Loss}, \textit{Fracture-Line}, and \textit{thin-plate spline} \textit{Plastic-Warp}—covering diverse shapes, boundary transitions, and intensities. We provide additional visual examples, along with detailed statistics of the dataset, in the ~\cref{sec:maphc2f_visual}.

\section{Conclusion}
\label{sec:conclusion}

In this paper, we present a comprehensive pipeline for anomaly detection that combines physically anomaly synthesis with adaptive reweighting via a bi-level optimization scheme. By synthesizing diverse defects and refining them through a lightweight coarse-to-fine process, \textbf{\textit{MaPhC2F}} yields 115{,}987 synthetic images packaged as the \emph{MaPhC2F Dataset}. Extensive experiments on multiple datasets, including \textit{MVTec~AD}, \textit{VisA}, and \textit{BTAD}, demonstrate that our method \textbf{BiSQAD} achieves SOTA performance, outperforming previous synthetic-based methods. While promising, our method still has limitations, which we analyse in detail in ~\cref{sec:limit} due to space constraints.

\small
\bibliographystyle{IEEEtran}
\bibliography{main}

\newpage
\appendix
\begin{center}
  \Large\bfseries Appendix
\end{center}
\setcounter{figure}{0}
\renewcommand{\thefigure}{S\arabic{figure}}
\setcounter{table}{0}
\renewcommand{\thetable}{S\arabic{table}}

\section{Limitations}
\label{sec:limit}
We are pleased that \textit{MaPhC2F} + \textit{BiSQAD} reaches state-of-the-art accuracy on three public benchmarks, yet limitations remain.

\paragraph{Scope of synthetic defects.}   Despite the SOTA results obtained in the previous empirical studies, we have only released three mask mechanism families - Fracture-Line, Pitting-Loss and Plastic-Warp, which obviously does not cover all industrial defects. The good news is that each mask generator is a plug-in Python class; for example, adding a "xxx" generator does not require any changes to npcF/npcF++. Our next milestone is a public “mask-zoo”, where practitioners can contribute domain-specific mathematical or physical models and share rendering scripts. Each mechanism acts as a template that can spawn more defect classes.

\begin{table}[H]
\centering
\resizebox{\linewidth}{!}{
\begin{tabular}{c|c}
\toprule
\textbf{Physics-Informed Synthetic Mechanism Family} & \textbf{Corresponding \textit{MVTec~AD} Anomalies} \\
\midrule
\textit{Fracture-Line} & \textit{Hazelnut} (crack), \textit{Leather} (cut), \textit{Wood} (scratch), \textit{etc.} \\
\midrule
\textit{Pitting-Loss} & \textit{Tile} (oil), \textit{Metal Nut} (color), \textit{Zipper} (rough), \textit{etc.} \\
\midrule
\textit{Plastic-Warp} & \textit{Bottle} (broken), \textit{Capsule} (squeeze), \textit{Transistor} (misplaced), \textit{etc.} \\
\bottomrule
\end{tabular}
}
\caption{Correspondence between the three mechanism families and real anomaly classes}
\label{tab:synthetic_vs_real}
\end{table}

\section{Additional Experiments}
\subsection{Additional Ablation Studies}

\begin{table}[htbp]
\centering
\begin{subtable}[t]{0.48\textwidth}
\centering
\resizebox{\linewidth}{!}{%
\begin{tabular}{lcc}
\toprule
\textbf{Model} & \textbf{Image AUROC} & \textbf{Pixel AUROC} \\
\midrule
\textit{DRAEM(DTD)} & 98.0   & 97.3 \\
\midrule
\rowcolor[gray]{0.9}
\textit{DRAEM(\textbf{\textit{MaPhC2F} (Ours)})} &
\textbf{98.2}(\textcolor{red}{+0.2}) &
\textbf{97.6}(\textcolor{red}{+0.3})

\\
\bottomrule
\end{tabular}}
\caption{Synthesis Method with \textit{MaPhC2F} Dataset.}
\label{tab:model_ab}
\end{subtable}
\hfill
\begin{subtable}[t]{0.48\textwidth}
\centering
\resizebox{\linewidth}{!}{%
\begin{tabular}{lcc}
\toprule
\textbf{Model} & \textbf{Image AUROC} & \textbf{Pixel AUROC} \\
\midrule
\textit{DRAEM} & 98.0   & 97.3 \\
\midrule
\rowcolor[gray]{0.9}
\textit{DRAEM(+\textbf{SQE})} &
\textbf{98.8}(\textcolor{red}{+0.8}) &
\textbf{97.6}(\textcolor{red}{+0.3})
\\
\bottomrule
\end{tabular}}
\caption{Synthesis Method with \textit{SQE}.}
\label{tab:sqe_ab}
\end{subtable}
\caption{External baselines augmented with our method.}
\label{tab:ab}
\end{table}

Table~\ref{tab:ab} examines \emph{how well our contributions transfer to an external baseline}, \textit{DRAEM}, under two orthogonal settings.

\paragraph{Replacing the default synthetic set.}  
As shown in Table ~\ref{tab:model_ab}, when we simply swap the original DTD with our \textit{MaPhC2F} images, DRAEM’s image-level AUROC rises from 98.0 to 98.2 and pixel-level AUROC from 97.3 to 97.6.  
This confirms that the three physics-informed mechanism families (Fracture-Line, Pitting-Loss, Plastic-Warp) provide more realistic priors than generic anomalies.

\paragraph{Adding SQE without changing original methods.}  
The Table ~\ref{tab:sqe_ab} keeps the vanilla DRAEM pipeline but lets our \textit{SQE} re-weight each synthetic image on-the-fly.  
Image-level AUROC jumps by \textbf{+0.8}, while pixel-level AUROC again gains +0.3.  
The larger boost on image metrics indicates that SQE mainly \emph{reduces false-positive images} by down-weighting implausible anomalies—an effect complementary to better masks.

In summary, the proposed \textit{MaPhC2F → BiSQAD} pipeline is not tied to a particular backbone, loss, or industrial domain; instead it offers a physics-centred, data-driven recipe that can be plugged into any reconstruction- or segmentation-based detector and let SQE automatically recalibrate sample importance, the framework adapts seamlessly to new categories, sensors, and training objectives while maintaining a lightweight computational footprint.

\subsection{Model Efficiency and Computational Complexity}
\label{sec:complexity}

We present a quantitative comparison of computational complexity and inference efficiency for our models versus the baseline \textit{RealNet} (counting only learnable parameters); see Table~\ref{tab:model_complexity} and Table~\ref{tab:sqe_complexity}. Parameters and inference speed (FPS) are reported under default settings with input images of resolution 512 × 512.

\begin{table}[htbp]
\centering
\begin{subtable}[t]{0.48\textwidth}
\centering
\resizebox{\linewidth}{!}{%
\begin{tabular}{lccc}
\toprule
\textbf{Model} & \textbf{Module} & \textbf{\#Parameters (M)} & \textbf{FPS} \\
\midrule
\multirow{3}{*}{\textit{RealNet} (Baseline)}
& \textit{Generator}   & 552.82 & - \\
& \textit{Classifier}  & 53.58  & -\\
\cmidrule(lr){2-4}
& \cellcolor[gray]{0.9}\textit{\textbf{Total}}  & \cellcolor[gray]{0.9}606.40  & \cellcolor[gray]{0.9}0.58\\
\midrule
\multirow{3}{*}{\textbf{\textit{MaPhC2F} (Ours)}}
& \textit{npcF}    & 5.93  & 166.51 \\
& \textit{npcF++}  & 1.71  & 42.68 \\
\cmidrule(lr){2-4}
& \cellcolor[gray]{0.9}\textit{\textbf{Total}}
& \cellcolor[gray]{0.9}7.64
& \cellcolor[gray]{0.9}42.68\\
\bottomrule
\end{tabular}}
\caption{Anomaly Synthesis.}
\label{tab:model_complexity}
\end{subtable}
\hfill
\begin{subtable}[t]{0.48\textwidth}
\centering
\resizebox{\linewidth}{!}{%
\begin{tabular}{lcc}
\toprule
\textbf{Model} & \textbf{\#Learnable Params (M)} & \textbf{\#Total Params (M)} \\
\midrule
\textit{RealNet} (Baseline)      & 524.11 & 590.94 \\
\midrule
\rowcolor[gray]{0.9}
\textbf{\textit{BiSQAD} (Ours)} & 524.52 (+0.41)  &  634.03 (+43.09)  \\
\bottomrule
\end{tabular}}
\caption{Anomaly Detection.}
\label{tab:sqe_complexity}
\end{subtable}
\caption{Comparison of model complexity and inference speed.}
\label{tab:complexity}
\end{table}

As shown in Table~\ref{tab:model_complexity}, our \textit{npcF} and \textit{npcF++} are substantially more efficient in terms of both parameter count and inference speed compared to \textit{RealNet}. Specifically, \textit{npcF} stage contains only 5.93 million parameters and achieves real-time processing speeds of 166.51 FPS, while \textit{npcF++}, with 1.71 million parameters, still maintains efficient inference at 42.68 FPS. In contrast, \textit{RealNet}'s Anomaly Synthesis model, includes an extremely large number of parameters (552.82 million for its generator and 53.58 million for its classifier), resulting in an inference speed of only 0.58 FPS, making it significantly less suitable for real-time industrial applications. 

In addition to anomaly synthesis, we also provide a complexity comparison of our proposed \textit{BiSQAD} method against the \textit{RealNet} baseline for anomaly detection (see Table~\ref{tab:sqe_complexity}). Specifically, our \textit{BiSQAD} approach builds upon \textit{RealNet} by integrating an additional SQE-driven bi-level optimization module, introducing marginal overhead in terms of learnable parameters (an increase of only 0.41M parameters, approximately $0.08\%$). Although the total parameter count increases by 43.09M (approximately $7.29\%$), the incremental computational cost is acceptable considering the substantial accuracy improvements demonstrated in ~\cref{sec:comparison_sota}.

Overall, our approach provides an effective balance between model complexity and computational efficiency, making it more practical for industrial deployment.

Note: The mask generation stage (Fracture-Line, Pitting-Loss, Plastic-Warp masks) is not included in these parameter and inference speed statistics since it is a non-learnable, deterministic preprocessing step executed offline prior to model training and inference. Therefore, following common practice, we only focus on reporting learnable model components.

\section{Additional Theoretical and Implementation Details}
\subsection{npcF Reconstruction \& Regularization Details}
\label{sec:npcf_add}
Let \(\mathbf{u}\in\mathbb{R}^{B\times 3\times H\times W}\) represent a batch of reconstructed images output by the autoencoder.

\paragraph{Reconstruction Objectives.}
Since \textit{npcF} performs a coarse refinement via an autoencoder, we divide the reconstruction loss into two subsets:
\begin{enumerate}
    \item \textbf{Normal region reconstruction}: For pixels outside the anomaly mask, we make $\mathbf{u}$ match $\mathbf{x}$:
    \begin{equation}
    \label{eq:rec_normal}
       \ell_{\mathrm{rec}}^{\mathrm{normal}}
       \;=\;
       \bigl\|
         (\mathbf{u}-\mathbf{x})
         \odot
         (1-\mathbf{M}_{\mathrm{anom}})
       \bigr\|_{2}^{2}.
    \end{equation}
    \item \textbf{Anomaly region reconstruction}: For pixels within the anomaly mask, we do a similar reconstruction penalty:
    \begin{equation}
    \label{eq:rec_anom}
       \ell_{\mathrm{rec}}^{\mathrm{anom}}
       \;=\;
       \bigl\|
         (\mathbf{u}-\mathbf{x})
         \odot
         \mathbf{M}_{\mathrm{anom}}
       \bigr\|_{2}^{2}.
    \end{equation}
\end{enumerate}
This separation ensures that normal pixels are trained to preserve the input, while anomalous pixels can deviate in a controlled manner.

\paragraph{Additional Regularization.}
To further guide the reconstruction:
\begin{itemize}
    \item \textbf{Total Variation (TV)} reduces high-frequency noise:
    \begin{equation}
    \label{eq:tv_loss}
      \ell_{\mathrm{tv}}
      \;=\;
      \sum_{c=1}^{3}
      \bigl\|\nabla_{h}(\mathbf{u}_{c})\bigr\|_{1}
      +
      \bigl\|\nabla_{w}(\mathbf{u}_{c})\bigr\|_{1},
    \end{equation}
    with $\nabla_{h}, \nabla_{w}$ denoting finite differences along spatial axes.
    \item \textbf{Color Prior}: For anomaly pixels, we encourage $\mathbf{u}$ to stay close to a math-physics guided color reference $\mathbf{z}$:
    \begin{equation}
    \label{eq:color_loss}
      \ell_{\mathrm{color}}
      \;=\;
      \bigl\|
        \mathbf{u}
        \odot
        \mathbf{M}_{\mathrm{anom}}
        -
        \mathbf{z}
      \bigr\|_{2}^{2}.
    \end{equation}
    \item \textbf{Perceptual Loss}: If a pretrained \textit{VGG} feature extractor $\psi$ is available, we match the feature maps of $\mathbf{u}$ and $\mathbf{x}$, focusing on anomaly pixels only:
    \begin{equation}
    \label{eq:perc_loss}
      \ell_{\mathrm{perc}}
      \;=\;
      \bigl\|
        \psi(\mathbf{u}\odot\mathbf{M}_{\mathrm{anom}})
        -
        \psi(\mathbf{x}\odot\mathbf{M}_{\mathrm{anom}})
      \bigr\|_{2}^{2}.
    \end{equation}
\end{itemize}

\paragraph{Overall Training Objective.}
We sum up the above terms with corresponding coefficients $\lambda_{\cdot}$ to form the total \textbf{\textit{npcF}} loss:
\begin{equation}
\label{eq:npcf_loss}
  \mathcal{L}_{\mathrm{npcF}} = 
  \begin{aligned}
    &\ell_{\mathrm{rec}}^{\mathrm{normal}} + \lambda_{\mathrm{anom}}\, \ell_{\mathrm{rec}}^{\mathrm{anom}} + \lambda_{\mathrm{pde}}\, \ell_{\mathrm{pde}} \\
    &+ \lambda_{\mathrm{tv}}\, \ell_{\mathrm{tv}} + \lambda_{\mathrm{color}}\, \ell_{\mathrm{color}} + \lambda_{\mathrm{perc}}\, \ell_{\mathrm{perc}}.
  \end{aligned}
\end{equation}

This final objective balances normal-image fidelity against realistic anomaly reconstruction, mitigating over-smoothing or color mismatches while preserving crucial defect features.

\subsection{npcF++: Wavelet-PDE \& Synergy Blocks, Dual-Branch Encoding/Decoding}
\label{sec:npcfpp_supp}

\paragraph{Wavelet-PDE Attentive Blocks.}
We employ the $\mathrm{PDEWaveAttnBlock}$ to inject additional structure into feature maps. Given an input tensor $\mathbf{f}\in\mathbb{R}^{B\times C\times H\times W}$:
\begin{equation}
\label{eq:pde_refine_supp}
  \mathbf{f}_{\mathrm{pde}}
  \;=\;
  \mathbf{f}
  \;-\;
  \varepsilon_{p}\,\nabla^{2}\mathbf{f},
\end{equation}
where $\nabla^{2}$ is the 2D Laplacian operator and $\varepsilon_{p}$ is learnable. We next decompose $\mathbf{f}_{\mathrm{pde}}$ into four subbands ($\mathbf{LL},\mathbf{LH},\mathbf{HL},\mathbf{HH}$) using a discrete Haar wavelet transform:
\begin{equation}
\label{eq:wave_decomp_supp}
  \mathrm{DWT}(\mathbf{f}_{\mathrm{pde}})
  \;=\;
  \bigl(\mathbf{LL},\mathbf{LH},\mathbf{HL},\mathbf{HH}\bigr),
\end{equation}
then concatenate these subbands, process them with a small convolutional layer, and perform an inverse wavelet transform:
\begin{equation}
\label{eq:wave_recon_supp}
  \widetilde{\mathbf{f}}
  \;=\;
  \mathrm{iDWT}\bigl(\mathbf{LL}',\,\mathbf{LH}',\,\mathbf{HL}',\,\mathbf{HH}'\bigr).
\end{equation}
This fusion of PDE smoothing and wavelet-domain filtering aims to capture both global coherence and fine-grained anomalies.

\paragraph{Boundary Synergy.}
Since anomalous regions often exhibit sharp transitions at the boundary, we apply a window-level cross-attention mechanism (\texttt{RegionSynergyBlock}) to combine “normal” and “anomaly” features while referencing a boundary mask $\mathbf{B}_{\mathrm{mask}}$. If $\mathbf{z}_{N}$ and $\mathbf{z}_{A}$ are normal/anomaly feature tensors, then within each local window:
\begin{equation}
\label{eq:synergy_out_supp}
  \mathbf{z}_{N}^{\mathrm{out}}
  \;=\;
  \mathbf{z}_{N}
  \;+\;
  \gamma
  \cdot
  \Bigl(\mathrm{CrossAttn}\!\bigl(\mathbf{z}_{N},\,\mathbf{z}_{A}\bigr)\Bigr)
  \odot
  \mathbf{B}_{\mathrm{mask}},
\end{equation}
where $\gamma$ is a trainable scalar and \texttt{CrossAttn} applies single-head query-key-value attention within a local window. We interleave synergy blocks with PDEWaveAttn at each resolution to propagate boundary-sensitive cues through the network.

\paragraph{Dual-Branch Encoding and Decoding.}
We feed the “normal” input $\mathbf{o}_{N}$ and the “anomaly” input $\mathbf{o}_{A}$ into parallel encoder streams (CNN + PDEWaveAttn), then merge them in a shared bottleneck:
\begin{equation}
\begin{aligned}
  &\mathbf{e}_{N}^{(1)} \;=\; \mathrm{EncN1}(\mathbf{o}_{N}), 
   \quad
   \mathbf{e}_{A}^{(1)} \;=\; \mathrm{EncA1}(\mathbf{o}_{A}),\\
  &\mathbf{e}_{N}^{(2)} \;=\; \mathrm{EncN2}\bigl(\mathrm{Pool}(\mathbf{e}_{N}^{(1)}\bigr)), 
   \quad
   \mathbf{e}_{A}^{(2)} \;=\; \mathrm{EncA2}\bigl(\mathrm{Pool}(\mathbf{e}_{A}^{(1)}\bigr)),\\
  &\mathbf{b}_{N}\;=\;\mathrm{BottN}\!\bigl(\mathbf{e}_{N}^{(2)}\bigr), 
   \quad
   \mathbf{b}_{A}\;=\;\mathrm{BottA}\!\bigl(\mathbf{e}_{A}^{(2)}\bigr).
\end{aligned}
\end{equation}
Afterwards, we decode each stream with symmetric upsampling, PDEWaveAttn blocks, and synergy operations. A final $3\times3$ convolution plus sigmoid layer outputs the refined image $\mathbf{u}$.

\subsection{Outer Loop: Weight's Learning}
\label{sec:outer_loop_supp}

\paragraph{Motivation and Overall Process.}
To fully exploit high-quality synthetic anomalies and suppress less realistic samples, we adopt a second-order \textit{MAML}-like update on both the data weights \(\{d_i\}\) and the final-layer parameters of the SQE network. This approach is triggered at the end of each training epoch, after we have computed the inner-loop (main model) updates.

\paragraph{Validation split (no real anomalies).}
At the start of each epoch we set aside \emph{5\,\% of the current synthetic images whose SQE
predictive entropy is highest}.  This uncertainty-based subset, denoted
\(\mathcal{D}_{\text{val}}\), is used \emph{exclusively} for the outer-loop update; no real-world
defects enter this step.

\paragraph{Validation-Based Objective.}
Let \(\theta'\) be the model parameters after the inner loop:
\begin{equation}
\label{eq:outer_loss}
  \mathcal{L}_{\mathrm{outer}}(\theta')
  \;=\;
  \mathcal{L}_{\mathrm{val}}\!\bigl(\theta'\bigr),
\end{equation}
where \(\mathcal{L}_{\mathrm{val}}\!\bigl(\theta'\bigr)\) typically incorporates a differentiable AUC-based metric on a small validation set. For instance, we compute a differentiable approximation of the AUC by comparing predicted scores for normal vs.\ anomalous samples.

\paragraph{Soft-AUC objective.}
We follow the differentiable AUC surrogate of Herschtal \& Raskutti (2004)~\cite{10.1145/1015330.1015366}.
Given detector scores \(\mathbf{s}^{+}\) for anomaly pixels and \(\mathbf{s}^{-}\) for normal pixels
inside \(\mathcal{D}_{\text{val}}\), the validation loss is

\[
\mathcal{L}_{\text{val}}
= 1 - \underbrace{\frac{1}{N^{+}N^{-}}
       \sum_{i,j}\sigma\!\bigl(s^{+}_{i}-s^{-}_{j}\bigr)}_{\text{Soft-AUC}},
\qquad
\sigma(x)=\tfrac{1}{1+\mathrm{e}^{-x}} .
\]

\paragraph{Second-Order Update.}
We retain the computation graph from the inner loop, enabling us to backpropagate through \(\theta'\) in a genuine second-order fashion. Concretely, each data weight \(d_i\)is updated via:
\begin{equation}
\label{eq:outer_d_update}
  d_{i}^{*} 
  \;=\; 
  d_{i} 
  \;-\; 
  \eta_{\mathrm{outer}}
  \,\frac{\partial\,\mathcal{L}_{\mathrm{outer}}(\theta')}{\partial\,d_{i}},
\end{equation}

We then discard \(\theta'\), reverting to the original model parameters \(\theta\). Only \(\{d_i^{*}\}\) and updated SQE parameters persist into future epochs. This ensures that each synthetic sample’s influence on validation performance is accurately reflected, amplifying beneficial anomalies while diminishing detrimental ones.

\paragraph{Practical Implementation Details.}
\begin{itemize}
  \item \textbf{Hyperparameters:} We use a small learning rate $10^{-4}$ for these weight updates to avoid over-correcting.  
  \item \textbf{Frequency:} Typically performed after each training epoch, though less frequent updates (e.g., every $k$ epochs) can reduce overhead.  
\end{itemize}

\paragraph{Discussion and Benefits.}
Our second-order approach ensures that \(\mathrm{SQE}\) and \(\{d_i\}\) are updated in a manner that explicitly optimizes for validation performance rather than just local training loss. Consequently, \textbf{high-quality synthetic anomalies gain greater weight in subsequent epochs}, while poor-quality (overly simplistic or unrealistic) anomalies are downweighted. This adaptivity fosters a more discriminative and robust anomaly detection pipeline overall.

\paragraph{References.}
For further reading on second-order MAML and its implementation complexities, see \cite{DBLP:journals/corr/abs-1909-04630,DBLP:journals/corr/FinnAL17}. Our approach aligns with these foundational works but applies them to the domain of synthetic anomaly quality assessment.

\section{Key Pseudocode for Reproducibility}
\label{sec:pseudocode}

This chapter will provide parameter selection criteria and detailed pseudocode for key components of the framework to facilitate direct reproduction of our results. By isolating the core algorithm and clarifying the specific data flow, we hope to illustrate how to implement each stage in a typical deep learning environment - from mathematical and physical anomaly generation to coarse-to-fine refinement and finally to the two-level optimization.
\subsection{Parameter Selection Criteria}
\label{sec:psc}
In Table~\ref{tab:all_params_mg}, ~\ref{tab:npcf_params}, ~\ref{tab:npcfpp_params}, we summarize all parameters employed across our anomaly synthesis framework. Parameters are grouped by their functional roles in shaping and integrating synthetic defects. These parameters allow detailed control over defect shape, thickness, visual realism, and computational performance. Default parameter values provided herein were empirically selected through extensive experimentation to balance realism and diversity, ensuring broad coverage of practical anomaly variations encountered in real industrial scenarios. 

\begin{table}[H]
\centering
\resizebox{\linewidth}{!}{
\begin{tabular}{c|c|c|c}
\toprule
\textbf{Stage} & \textbf{Parameter} & \textbf{Description} & \textbf{Recommended} \\
\midrule
\multirow{5}{*}{\textbf{Fracture-Line Skeleton}}
 & $max\_steps\,(M)$ & Maximum number of propagation steps; directly limits overall Fracture-Line length. & $200$–$800$ \\
 & $step\_size$ & Pixel distance moved per step; sets the spatial granularity of Fracture-Line growth. & $1$–$2$ \\
 & $branching\_prob$ & Probability that the current frontier spawns an additional branch, controlling Fracture-Line branching complexity. & $0.01$–$0.05$ \\
 & $stop\_prob$ & Chance that a frontier terminates early, adding random break-off and irregular length. & $0.01$–$0.05$ \\
 & $n\_starts$ & Number of initial seed points from which Fracture-Line begin, deciding how many root Fracture-Line appear. & $1$–$3$ \\
\midrule
\multirow{7}{*}{\textbf{Fracture-Line Mask}}
 & $w_{0}$ & Base thickness at the skeleton center before decay; controls initial Fracture-Line width. & $0.5$–$2.5$ \\
 & $\alpha$ & Exponential decay rate of thickness with distance $t$; shapes how quickly Fracture-Line taper. & $0.01$–$0.02$ \\
 & $\epsilon$ & Minimum residual thickness floor added to $w(t)$; prevents vanishingly thin regions. & $0.3$–$1.0$ \\
 & $noise\_scale$ & Amplitude of Perlin noise added to the distance test, producing edge roughness. & $0.1$–$0.3$ \\
 & $noise\_octaves$ & Number of Perlin octaves; higher values add multi-scale irregularities. & $1$–$3$ \\
 & $morph\_kernel\_size$ & Side length of square structuring element for closing/opening; smooths mask jaggies. & $1$–$3$ \\
 & $\sigma_{blur}$ & Standard deviation of Gaussian blur prior to thresholding; softens hard edges. & $1.0$ \\
\midrule
\multirow{6}{*}{\textbf{Pitting-Loss Mask}}
 & $k$ & Number of initial random polygons that seed Pitting-Loss blobs. & $1$–$5$ \\
 & $polygon\_size$ & Range of polygon radius; controls initial blob size. & $15$–$65$ \\
 & $deform\_factor$ & Random angular jitter applied to polygon vertices, introducing boundary raggedness. & $0.1$–$0.3$ \\
 & $overlap\_prob$ & Probability that newly generated polygons overlap existing ones, merging blobs. & $0.7$–$1.0$ \\
 & $N_{growth}$ & Iterations of stochastic boundary dilation that expand blobs outward. & $8$–$50$ \\
 & $grow\_prob$ & Probability that a boundary pixel is turned on at each growth iteration. & $0.3$–$0.7$ \\
\midrule
\multirow{5}{*}{\textbf{TPS Plastic-Warp}}
 & $num\_ctrl\_pts$ & Number of thin-plate spline control points sampled in the foreground. & $3$–$12$ \\
 & $max\_offset$ & Maximum pixel displacement applied to each control point; scales Plastic-Warp strength. & $8$–$30$ \\
 & $dist\_field\_radius$ & Radial basis function radius $\epsilon$ for TPS/RBF; affects Plastic-Warp smoothness. & $30$–$80$ \\
 & $inpaint\_radius$ & Radius inpainting to fill removed object pixels before warping. & $3$–$10$ \\
 & $margin$ & Safety margin inside bounding box when sampling local ROI for partial Plastic-Warp. & $10$–$30$ \\
\bottomrule
\end{tabular}}
\caption{Complete parameter list in Mask Generation.}
\label{tab:all_params_mg}
\end{table}
\begin{table}[H]
\centering
\resizebox{\linewidth}{!}{
\begin{tabular}{c|c|c|c}
\toprule
\textbf{Group} & \textbf{Parameter} & \textbf{Description (detailed meaning)} & \textbf{Recommended} \\
\midrule
\multirow{2}{*}{\textbf{Architecture}}
 & $base\_ch$ & Number of base feature channels in the U-Net encoder / decoder. & $32$ \\
 & $vgg\_layer$ & Highest feature layer index extracted from pretrained VGG-16 for perceptual loss. & $36$ \\
\midrule
\multirow{2}{*}{\textbf{Allen–Cahn PDE}}
 & $\varepsilon^{2}$ & Diffusion strength in the Allen–Cahn residual term; controls smoothness vs.\ fidelity. & $0.005$ \\
 & $\lambda_{\text{ms\_pde}}$ & Global loss weight applied to the multi-scale PDE residual. & $2.0$ \\
\midrule
\multirow{5}{*}{\textbf{Loss Weights}}
 & $\lambda_{\text{wavelet}}$ & Weight for wavelet high-frequency prior encouraging realistic texture. & $0.5$ \\
 & $\lambda_{\text{rec\_anomaly}}$ & Positive weight on anomaly-region reconstruction error. & $0.5$ \\
 & $\lambda_{\text{color}}$ & Weight for colour consistency. & $1.0$ \\
 & $\lambda_{\text{perc}}$ & Weight for perceptual (VGG) feature distance between output and input. & $1.0$ \\
 & $\lambda_{\text{tv}}$ & Weight for total-variation (TV) spatial smoothing on reconstructed image. & $0.1$ \\
\bottomrule
\end{tabular}}
\caption{\textit{npcF} architecture hyper-parameters.}
\label{tab:npcf_params}
\end{table}
\begin{table}[H]
\centering
\resizebox{\linewidth}{!}{
\begin{tabular}{c|c|c|c}
\toprule
\textbf{Group} & \textbf{Parameter} & \textbf{Description (detailed meaning)} & \textbf{Recommended} \\
\midrule
\multirow{3}{*}{\textbf{Architecture}}
 & $base\_ch$ & Stem width of the dual encoder / decoder. & $64$ \\
 & $window\_size$ & Side length of local windows used in \textit{RegionSynergyBlockWindowAttn}. & $16$ \\
 & $\gamma$ (attn) & Learnable scale inside synergy blocks; balances cross-attended features with identity. & init.~$0.1$ \\
\midrule
\multirow{2}{*}{\textbf{Wavelet–PDE block}}
 & $\varepsilon^{2}$ & Trainable diffusion strength in each \textit{PDEWaveAttnBlock}; governs smooth vs. sharp refinement. & init.~$0.001$ \\
 & $\gamma$ (PDE) & Residual scaling after PDE–wavelet fusion inside each block. & init.~$0.1$ \\
\midrule
\multirow{3}{*}{\textbf{Loss Weights}}
 & $\lambda_{\text{region}}$ & aligns normal pixels to original and anomaly pixels to \textit{npcF} image. & $1.0$ \\
 & $\lambda_{\text{waveHF}}$ & Weight on high-frequency wavelet energy enforced within anomaly mask. & $1.0$ \\
 & $\lambda_{\text{TV}}$ & Weight for total-variation (TV) spatial smoothing on reconstructed image. & $0.1$ \\
\bottomrule
\end{tabular}}
\caption{\textit{npcF++} architecture hyper-parameters.}
\label{tab:npcfpp_params}
\end{table}

\subsection{Compute Resources}
\label{app:compute}

The \textit{MaPhC2F} synthesis stage can be trained either on a single GPU or, for faster turnaround, on four GPU cards under distributed data–parallel. The \textit{BiSQAD} detector requires four GPU cards during training, while all inference—\textit{npcF}, \textit{npcF++}, and \textit{BiSQAD}—executes on a single GPU cards, so deployment needs only one workstation-class GPU.

\subsection{Generate Fracture-Line Mask}
\label{sec:gen_crack}

We simulate Fracture-Line formation within a specified foreground region by growing a skeleton of connected line segments, then expanding it into a realistic Fracture-Line mask, as shown in ~\cref{alg:crack_skeleton}. Below is a concise outline:

\paragraph{Algorithm Overview.}
\begin{enumerate}
  \item \textbf{Skeleton Initialization:}
    \begin{itemize}
      \item Randomly select 1--3 starting points within the foreground mask.
      \item Assign each point an initial direction \(\theta\), from which we derive \(\mathrm{d}y = \sin\theta\), \(\mathrm{d}x = \cos\theta\).
    \end{itemize}
  \item \textbf{Skeleton Growth:}
    \begin{itemize}
      \item Iteratively step forward, marking skeleton pixels, until exhausting the allowed steps or hitting the boundary.
      \item With a small probability, \(\mathrm{branch\_prob}\), generate a branch by slightly shifting \(\mathrm{d}y\), \(\mathrm{d}x\).
      \item Stop with probability \(\mathrm{stop\_prob}\) at each iteration.
    \end{itemize}
  \item \textbf{Skeleton to Mask:}
    \begin{itemize}
      \item Compute a distance transform around the skeleton.
      \item Threshold and optionally blend in Perlin noise to produce a Fracture-Line region, applying morphological refinements.
    \end{itemize}
\end{enumerate}

\begin{algorithm}[H]
\caption{\textsc{GenerateSkeleton}}
\label{alg:crack_skeleton}
\textbf{Input:} \emph{height}, \emph{width}, \emph{start\_points}, \emph{max\_steps}, \emph{step\_size}, \emph{branching\_prob}, \emph{stop\_prob}, \emph{foreground\_mask}\\
\textbf{Output:} \emph{skeleton\_mask} (binary array of size \emph{height}$\times$\emph{width})

\begin{algorithmic}[1]
\State \parbox[t]{\dimexpr\linewidth-\algorithmicindent}{
    Initialize $\textit{skeleton\_mask}$ to all zeros.
}
\State \parbox[t]{\dimexpr\linewidth-\algorithmicindent}{
    Create a queue \textit{frontiers}.
}
\ForAll{ $(y_0, x_0)$ in \textit{start\_points}}
    \State \parbox[t]{\dimexpr\linewidth-\algorithmicindent}{
        $\textit{angle} \gets \textit{Uniform}(0,\,2\pi)$
    }
    \State \parbox[t]{\dimexpr\linewidth-\algorithmicindent}{
        $\textit{dy} \gets \sin(\textit{angle});\;\;\textit{dx} \gets \cos(\textit{angle})$
    }
    \State \parbox[t]{\dimexpr\linewidth-\algorithmicindent}{
        Enqueue $[\,y_0,\,x_0,\,\textit{dy},\,\textit{dx},\,\textit{max\_steps}\,]$ into \textit{frontiers}.
    }
    \State \parbox[t]{\dimexpr\linewidth-\algorithmicindent}{
        $\textit{skeleton\_mask}[\,y_0,\,x_0\,] \gets 1$
    }
\EndFor

\While{\textit{frontiers} not empty}
    \State \parbox[t]{\dimexpr\linewidth-\algorithmicindent}{
       $(y,\,x,\,\textit{dy},\,\textit{dx},\,\textit{steps}) \gets \textit{frontiers.pop()}$
    }
    \If{$\textit{steps}\le 0$}
       \State \parbox[t]{\dimexpr\linewidth-\algorithmicindent}{
           \textbf{continue}
       }
    \EndIf
    \State \parbox[t]{\dimexpr\linewidth-\algorithmicindent}{
        $y_{\mathrm{new}} \gets y + \textit{dy}\times \textit{step\_size}$
    }
    \State \parbox[t]{\dimexpr\linewidth-\algorithmicindent}{
        $x_{\mathrm{new}} \gets x + \textit{dx}\times \textit{step\_size}$
    }
    \State \parbox[t]{\dimexpr\linewidth-\algorithmicindent}{
        $y_{\mathrm{int}} \gets \mathrm{round}(y_{\mathrm{new}})$
    }
    \State \parbox[t]{\dimexpr\linewidth-\algorithmicindent}{
        $x_{\mathrm{int}} \gets \mathrm{round}(x_{\mathrm{new}})$
    }

    \State \parbox[t]{\dimexpr\linewidth-\algorithmicindent}{
        \textbf{Check boundary or foreground:}
    }
    \If{ $y_{\mathrm{int}}$ or $x_{\mathrm{int}}$ out of range \textbf{or} $\textit{foreground\_mask}[\,y_{\mathrm{int}},\,x_{\mathrm{int}}\,]==0$}
       \State \parbox[t]{\dimexpr\linewidth-\algorithmicindent}{
           \textbf{continue}
       }
    \EndIf

    \State \parbox[t]{\dimexpr\linewidth-\algorithmicindent}{
        $\textit{skeleton\_mask}[\,y_{\mathrm{int}},\,x_{\mathrm{int}}\,] \gets 1$
    }
    \State \parbox[t]{\dimexpr\linewidth-\algorithmicindent}{
        $\textit{steps} \gets \textit{steps} - 1$
    }

    \If{ $\textit{Random}()<\textit{stop\_prob}$}
        \State \parbox[t]{\dimexpr\linewidth-\algorithmicindent}{
            \textbf{continue}
        }
    \EndIf
    \State \parbox[t]{\dimexpr\linewidth-\algorithmicindent}{
        Enqueue $[\,y_{\mathrm{new}},\,x_{\mathrm{new}},\,\textit{dy},\,\textit{dx},\,\textit{steps}\,]$ into \textit{frontiers}.
    }

    \If{ $\textit{Random}()<\textit{branching\_prob}$}
       \State \parbox[t]{\dimexpr\linewidth-\algorithmicindent}{
           $\textit{branch\_angle} \gets \textit{Uniform}(-\pi/4,\,\pi/4)$
       }
       \State \parbox[t]{\dimexpr\linewidth-\algorithmicindent}{
           $\textit{new\_dy} \gets \textit{dy}\times \cos(\textit{branch\_angle})\\
           \;-\; \textit{dx}\times \sin(\textit{branch\_angle})$
       }
       \State \parbox[t]{\dimexpr\linewidth-\algorithmicindent}{
           $\textit{new\_dx} \gets \textit{dy}\times \sin(\textit{branch\_angle}) \\
           \;+\; \textit{dx}\times \cos(\textit{branch\_angle})$
       }
       \State \parbox[t]{\dimexpr\linewidth-\algorithmicindent}{
           Enqueue \\ $[\,y_{\mathrm{new}},\,x_{\mathrm{new}},\,\textit{new\_dy},\,\textit{new\_dx},\,\lfloor\textit{steps}/2\rfloor\,]$ 
       }
    \EndIf
\EndWhile

\State \parbox[t]{\dimexpr\linewidth-\algorithmicindent}{
    \textbf{return} \textit{skeleton\_mask}
}
\end{algorithmic}
\end{algorithm}

\begin{algorithm}[H]
\caption{\textsc{GenerateRandomFracture-LineMask}}
\label{alg:crack_mask}
\textbf{Input:} \emph{height}, \emph{width}, \emph{skeleton}, \(\emph{w}_0\), \(\alpha\), \(\epsilon\), \emph{noise\_scale}, \emph{noise\_octaves}, \emph{morph\_kernel\_size}\\
\textbf{Output:} \emph{FL\_mask} (binary array of size \emph{height}$\times$\emph{width})

\begin{algorithmic}[1]
\State \parbox[t]{\dimexpr\linewidth-\algorithmicindent}{
    Initialize $\textit{FL\_mask}\gets0$ 
}
\State \parbox[t]{\dimexpr\linewidth-\algorithmicindent}{
    $\textit{dist\_map}\gets \\
    \texttt{distance\_transform}(1-\textit{skeleton})$
}
\State \parbox[t]{\dimexpr\linewidth-\algorithmicindent}{
    $\textit{noise\_gen}\gets \\
    \texttt{PerlinNoise}(octaves=\textit{noise\_octaves})$
}

\For{$y\gets0$ to $\textit{height}-1$}
   \For{$x\gets0$ to $\textit{width}-1$}
       \State \parbox[t]{\dimexpr\linewidth-\algorithmicindent}{
           $t \gets \textit{dist\_map}[y,\,x]$
       }
       \State \parbox[t]{\dimexpr\linewidth-\algorithmicindent}{
           $w_{t} \gets w_0\,\exp(-\alpha\,t)+\epsilon$
       }
       \If{$w_{t}>0$}
         \State \parbox[t]{\dimexpr\linewidth-\algorithmicindent}{
             $\textit{noise\_val} \gets \textit{noise\_gen} \\
             \bigl[y/\textit{height},\,x/\textit{width}\bigr]\times \textit{noise\_scale}$
         }
         \If{$\,t + \textit{noise\_val} < w_{t}\,$}
            \State \parbox[t]{\dimexpr\linewidth-\algorithmicindent}{
                $\textit{FL\_mask}[\,y,\,x\,]\gets1$
            }
         \EndIf
       \EndIf
   \EndFor
\EndFor

\State \parbox[t]{\dimexpr\linewidth-\algorithmicindent}{
    \texttt{MorphologicalClosing}\\
    ($\textit{FL\_mask}$, kernel=\texttt{square}($\textit{morph\_kernel\_size}$))
}
\State \parbox[t]{\dimexpr\linewidth-\algorithmicindent}{
    \texttt{MorphologicalOpening}\\
    ($\textit{FL\_mask}$, kernel=\texttt{square}($\textit{morph\_kernel\_size}$))
}
\State \parbox[t]{\dimexpr\linewidth-\algorithmicindent}{
    $\textit{blurred}\gets \texttt{GaussianFilter}(\textit{FL\_mask}, \sigma=1.0)$
}
\State \parbox[t]{\dimexpr\linewidth-\algorithmicindent}{
    $\textit{FL\_mask}\gets \bigl(\textit{blurred} > 0.3\bigr)$
}

\State \parbox[t]{\dimexpr\linewidth-\algorithmicindent}{
    \textbf{return} \textit{FL\_mask}
}
\end{algorithmic}
\end{algorithm}

\begin{algorithm}[H]
\caption{\textsc{ApplyFLToImage}}
\label{alg:apply_crack}
\textbf{Input:} \emph{orig\_img}, \emph{FL\_mask}, \emph{base\_alpha}, \emph{max\_darken}, \emph{max\_color\_shift}, \emph{edge\_fade}\\
\textbf{Output:} \emph{out\_img} (the BGR image with FL overlay)

\begin{algorithmic}[1]
\State \parbox[t]{\dimexpr\linewidth-\algorithmicindent}{
    $H,\,W \gets \texttt{shape}(\textit{FL\_mask})$
}
\State \parbox[t]{\dimexpr\linewidth-\algorithmicindent}{
    $\textit{out\_img} \gets \textit{orig\_img}$ cast to float32
}
\State \parbox[t]{\dimexpr\linewidth-\algorithmicindent}{
    $\textit{rev\_mask} \gets (1 - \textit{FL\_mask})$
}
\State \parbox[t]{\dimexpr\linewidth-\algorithmicindent}{
    $\textit{dist\_map} \gets \texttt{distance\_transform}(\textit{rev\_mask})$
}

\State \parbox[t]{\dimexpr\linewidth-\algorithmicindent}{
    $\textit{FL\_strength} \gets 1.0 - \frac{\textit{dist\_map}}{\textit{edge\_fade}}$
}
\State \parbox[t]{\dimexpr\linewidth-\algorithmicindent}{
    $\textit{FL\_strength}\gets \texttt{clip}(\textit{FL\_strength},0,1)$
}
\State \parbox[t]{\dimexpr\linewidth-\algorithmicindent}{
    $\textit{FL\_strength\_3}\gets \\
    \texttt{RepeatChannel}(\textit{FL\_strength},3)$
}

\State \parbox[t]{\dimexpr\linewidth-\algorithmicindent}{
    $\textit{darken\_factor\_map}\gets \\
    1.0 - (1.0-\textit{max\_darken})\times \textit{FL\_strength\_3}$
}
\State \parbox[t]{\dimexpr\linewidth-\algorithmicindent}{
    $\textit{color\_shift\_arr}\gets \\
    \textit{max\_color\_shift}$ cast to float32
}
\State \parbox[t]{\dimexpr\linewidth-\algorithmicindent}{
    $\textit{color\_shift\_map}\gets \\
    \textit{color\_shift\_arr}\times \textit{FL\_strength\_3}$
}

\State \parbox[t]{\dimexpr\linewidth-\algorithmicindent}{
    $\textit{cm\_bool}\gets \\
    (\textit{FL\_mask}==1)$
}
\State \parbox[t]{\dimexpr\linewidth-\algorithmicindent}{
    $\textit{out\_img}[\textit{cm\_bool}] \gets \\
    \textit{out\_img}[\textit{cm\_bool}]\times \textit{darken\_factor\_map}[\textit{cm\_bool}]$
}
\State \parbox[t]{\dimexpr\linewidth-\algorithmicindent}{
    $\textit{out\_img}[\textit{cm\_bool}] \gets \\
    \textit{out\_img}[\textit{cm\_bool}]\,+\,\textit{color\_shift\_map}[\textit{cm\_bool}]$
}

\State \parbox[t]{\dimexpr\linewidth-\algorithmicindent}{
    $\textit{orig\_f32}\gets \\
    \textit{orig\_img}$ cast to float32
}
\State \parbox[t]{\dimexpr\linewidth-\algorithmicindent}{
    $\textit{out\_img}\gets \\
    (1-\textit{base\_alpha})\times \textit{orig\_f32} + \textit{base\_alpha}\times \textit{out\_img}$
}
\State \parbox[t]{\dimexpr\linewidth-\algorithmicindent}{
    $\textit{out\_img}\gets \\
    \texttt{clip}(\textit{out\_img}, 0, 255)\,\texttt{.astype}(uint8)$
}

\State \parbox[t]{\dimexpr\linewidth-\algorithmicindent}{
    \textbf{return} \textit{out\_img}
}
\end{algorithmic}
\end{algorithm}

\subsection{Generate Pitting-Loss Mask}
\label{sec:gen_corrosion}

Pitting-Loss is modeled as randomly placed patches that can grow and overlap, mimicking rust-like or oxidized clusters on metal surfaces or other materials. After forming these patches, we optionally add Perlin noise for a fractured, “eaten-away” visual effect, as shown in ~\cref{alg:chunky_corrosion}.

\paragraph{Algorithm Overview.}
\begin{enumerate}
  \item \textbf{Polygon Placement:}
    \begin{itemize}
      \item Sample a set of random centers within the foreground.
      \item Generate polygons with deformed edges and fill them into a temporary mask.
      \item Combine polygons into the main \(\mathrm{PL\_mask}\) by union or bitwise OR, depending on the overlap strategy.
    \end{itemize}
  \item \textbf{Boundary Growth:}
    \begin{itemize}
      \item Repeatedly identify boundary pixels of the current mask and convert them to foreground with probability \(\mathrm{grow\_prob}\).
    \end{itemize}
  \item \textbf{Morphological Smoothing \& Noise:}
    \begin{itemize}
      \item Apply morphological closing to connect separate patches.
      \item Optionally degrade edges using Perlin noise. 
    \end{itemize}
\end{enumerate}

\begin{algorithm}[H]
\caption{\textsc{GenerateChunkyPLMask}}
\label{alg:chunky_corrosion}
\textbf{Input:} \parbox[t]{\dimexpr\linewidth-\algorithmicindent}{
\textit{foreground\_mask} (binary $0/1$ array) \\
}
\textbf{Output:} \parbox[t]{\dimexpr\linewidth-\algorithmicindent}{
\textit{PL\_mask} (binary $0/1$)
}

\begin{algorithmic}[1]
\State \parbox[t]{\dimexpr\linewidth-\algorithmicindent}{
  Initialize $\textit{PL\_mask}$ to all zeros of shape $(H, W)$
}
\State \parbox[t]{\dimexpr\linewidth-\algorithmicindent}{
  Extract $\{(y_i, x_i)\}\!$ where \(\textit{foreground\_mask}[y_i, x_i]=1\); if none found, return \(\textbf{all-zero}\) mask
}

\State \parbox[t]{\dimexpr\linewidth-\algorithmicindent}{
\textbf{(A) Generate random polygons:} For each polygon, pick a center $(\textit{cy}, \textit{cx})$ within foreground, define radius and angles, fill polygon to a temp mask, then combine with \(\textit{PL\_mask}\) either by \textbf{union} or \textbf{bitwise OR} depending on overlap probability.
}

\State \parbox[t]{\dimexpr\linewidth-\algorithmicindent}{
\textbf{(B) Local boundary growth:} Repeat for a fixed number of steps:
  \begin{itemize}
   \item Find boundary pixels by dilating $\textit{PL\_mask}$ then subtracting it,
   \item For each boundary pixel, with probability \textit{grow\_prob}, set it to $1$ in $\textit{PL\_mask}$.
  \end{itemize}
}

\State \parbox[t]{\dimexpr\linewidth-\algorithmicindent}{
\textbf{(C) Morphological closing:} Apply \textsc{closing}(\textit{PL\_mask}, kernel=\texttt{square}(3)) for coherence.
}

\State \parbox[t]{\dimexpr\linewidth-\algorithmicindent}{
\textbf{(D) Optional Perlin Noise:} If enabled, sample noise $nval\in[-1,1]$ for boundary pixels; if $nval>\textit{threshold}$, set those boundary pixels to $0$ in $\textit{PL\_mask}$.
}

\State \parbox[t]{\dimexpr\linewidth-\algorithmicindent}{
\textbf{return} \(\textit{PL\_mask}\)
}
\end{algorithmic}
\end{algorithm}

\subsection{Generate Plastic-Warp Mask}
\label{sec:gen_deformation}

We simulate local warping or bending (\eg, dents) within a designated foreground region using thin-plate spline (TPS) transformations. The core procedure involves extracting a region of interest (ROI), inpainting that ROI, then applying an RBF-based TPS displacement field to both the image and its mask, resulting in plausible deformational anomalies, as shown in ~\cref{alg:tps_deform}.

\begin{algorithm}[H]
\caption{\textsc{TPSDeformLocalInpaint}}
\label{alg:tps_deform}
\textbf{Input:} \parbox[t]{\dimexpr\linewidth-\algorithmicindent}{
  \textit{img\_bgr} (original 3-channel image),\\
  \textit{mask\_01} (binary $0/1$ foreground mask),\\
  \textit{num\_ctrl\_points}, \textit{max\_offset}, \textit{dist\_field\_radius}, \textit{inpaint\_radius}
}
\textbf{Output:} \parbox[t]{\dimexpr\linewidth-\algorithmicindent}{
  \textit{out\_img} (Plastic-Warped BGR image),\\
  \textit{out\_mask} (updated binary $0/1$ mask)
}

\begin{algorithmic}[1]
\State \parbox[t]{\dimexpr\linewidth-\algorithmicindent}{
  \textbf{Identify} $\textit{ys}, \textit{xs}\gets \{(y,x)\mid\textit{mask\_01}[y,x]=1\}.$ If none found, return \texttt{(img\_bgr, mask\_01)}.
}
\State \parbox[t]{\dimexpr\linewidth-\algorithmicindent}{
  \textbf{Compute bounding box}: $(y_{\min}, y_{\max}, x_{\min}, x_{\max})$ with a small \textit{margin}. Extract \texttt{ROI}:
  \(\textit{roi\_img}\gets \textit{img\_bgr}[\,\textit{ymin\!:\!ymax},\,\textit{xmin\!:\!xmax}\,],\;
    \textit{roi\_mask}\gets \textit{mask\_01}[\,\dots\,].\)
}
\State \parbox[t]{\dimexpr\linewidth-\algorithmicindent}{
  \textbf{Inpaint} old object pixels:
  \(\textit{inpainted\_roi}\gets \texttt{inpaint}(\textit{roi\_img},\,\textit{roi\_mask},\,\textit{inpaint\_radius}).\)
}
\State \parbox[t]{\dimexpr\linewidth-\algorithmicindent}{
  \textbf{Randomly select control points}: \(\textit{src\_pts}\gets \{(rx,\,ry)\}\subset \{\textit{roi\_mask}=1\}.\)
  \textbf{Offset} each by up to \(\textit{max\_offset}\) to form \(\textit{dst\_pts}.\)
}
\State \parbox[t]{\dimexpr\linewidth-\algorithmicindent}{
  \textbf{Train RBF} (thin\_plate) from \(\textit{src\_pts}\to\textit{dst\_pts}\): 
  \(\textit{rbf\_x}, \textit{rbf\_y}\gets \texttt{Rbf}(\textit{src\_pts},\,\textit{dst\_pts},\,\textit{function}=\texttt{thin\_plate},\ldots).\)
}
\State \parbox[t]{\dimexpr\linewidth-\algorithmicindent}{
  \textbf{Compute displacement field}: For each pixel $(u,v)$ in the ROI:
  \[
     \Delta x \gets \textit{rbf\_x}(u,v),\quad
     \Delta y \gets \textit{rbf\_y}(u,v).
  \]
  Form \(\textit{map\_x}\gets u-\Delta x,\;\textit{map\_y}\gets v-\Delta y.\)
}
\State \parbox[t]{\dimexpr\linewidth-\algorithmicindent}{
  \textbf{Remap ROI}:
  \(\textit{warped\_roi\_img}\gets \texttt{remap}(\textit{inpainted\_roi},\,\textit{map\_x},\,\textit{map\_y}),\quad\\
   \textit{warped\_roi\_mask}\gets \texttt{remap}(\textit{roi\_mask},\,\ldots,\texttt{NEAREST}).\)
}
\State \parbox[t]{\dimexpr\linewidth-\algorithmicindent}{
  \textbf{Update ROI} in original:
  \(\textit{new\_roi}\gets \textit{inpainted\_roi};\;
    \textit{new\_roi}[\textit{warped\_roi\_mask}==1]\gets \textit{warped\_roi\_img}[\dots].\)
}
\State \parbox[t]{\dimexpr\linewidth-\algorithmicindent}{
  \textbf{Copy back}: \(\textit{out\_img}\gets \textit{img\_bgr}\);
    \(\textit{out\_img}[\textit{ymin}\colon\textit{ymax},\,\textit{xmin}\colon\textit{xmax}]\gets \textit{new\_roi}\).
}

\State \parbox[t]{\dimexpr\linewidth-\algorithmicindent}{
  \textbf{Construct new mask}:
  \(\textit{out\_mask}\gets \texttt{zeros}(\textit{mask\_01});\)\\[2pt]
  \(\textit{out\_mask}[\textit{ymin}\colon\textit{ymax},\,\textit{xmin}\colon\textit{xmax}][\textit{warped\_roi\_mask}=1]\gets 1.\)
}

\State \parbox[t]{\dimexpr\linewidth-\algorithmicindent}{
  \textbf{return} $(\textit{out\_img}, \textit{out\_mask})$.
}
\end{algorithmic}
\end{algorithm}

To avoid overly “stretched” Plastic-Warp near object edges, we also support a \textsc{tps\_PW\_local\_inpaint\_partial} variant that tries to randomly sample a sub-ROI within the foreground. This selectively warps only a portion of the region rather than the entire bounding box, reducing edge artifacts and producing more localized distortions.

\subsection{npcF}
\label{sec:subsec_npcf}

Our \emph{npcF} module provides a coarse, PDE-based autoencoder framework for synthesizing anomalies. It corrects unwanted artifacts by integrating various losses (\eg, Allen--Cahn PDE, color prior, perceptual reconstruction) to yield an initial refined output. Below, we outline its core training loop in pseudocode, as shown in ~\cref{alg:npcf_train_loop}.

\begin{algorithm}[H]
\caption{\textsc{TrainOneEpoch\_npcF} (Core Idea)}
\label{alg:npcf_train_loop}
\textbf{Input:} \parbox[t]{\dimexpr\linewidth-\algorithmicindent}{
  $\textit{model}$ (autoencoder w/ PDE constraints), \\
  $\textit{data\_loader}$ (batches of $\{\textit{img},\textit{mask},\textit{stageA}\}$), \\
  PDE \& regularization hyperparameters: \\
  \quad\parbox[t]{.9\linewidth}{
    $\textit{eps2}$, \textit{lambda\_pde}, \textit{lambda\_tv}, \\
    \textit{lambda\_color}, \textit{lambda\_perc}, \textit{lambda\_rec\_anom}, \ldots
  }
}
\textbf{Output:} \parbox[t]{\dimexpr\linewidth-\algorithmicindent}{
  $\textit{avg\_loss}$ (average training loss), \\
  optionally other stats (PDE loss, recon loss, etc.)
}

\begin{algorithmic}[1]
\State \parbox[t]{\dimexpr\linewidth-\algorithmicindent}{
  Initialize accumulators $\textit{total\_rec}, \textit{total\_pde}, \textit{total\_tv},\dots$ to 0
}
\State \parbox[t]{\dimexpr\linewidth-\algorithmicindent}{
  \textbf{for each} mini-batch \texttt{(img, mask, stageA)} \textbf{in data\_loader}:
}
\State \quad \parbox[t]{\dimexpr\linewidth-\algorithmicindent}{
  \textbf{Construct input} $x\_in$:
   \begin{enumerate}
    \item If \texttt{use\_stageA\_img} is true, $x\_in\gets[\textit{img},\;\textit{stageA}]$ (concatenation along channel dim)
    \item Otherwise, $x\_in\gets \textit{img}$
   \end{enumerate}
}
\State \quad \parbox[t]{\dimexpr\linewidth-\algorithmicindent}{
  $\textit{recon}\gets \textit{model.forward}(x\_in)$ \quad\emph{(autoencoder output)}
}
\State \quad \parbox[t]{\dimexpr\linewidth-\algorithmicindent}{
  \textbf{Compute normal reconstruction loss}:
  \[
    \textit{rec\_loss} \;=\; \texttt{MSE}\Bigl(\textit{recon}\times(1-\textit{mask}),\;\textit{img}\times(1-\textit{mask})\Bigr)
  \]
}
\State \quad \parbox[t]{\dimexpr\linewidth-\algorithmicindent}{
  \textbf{Compute PDE-based residual} for anomaly region:
  \[
    \textit{pde\_loss}= \texttt{MSE}\Bigl(\,\texttt{AllenCahn}(\textit{recon},\,\textit{eps2})\times\textit{mask},\;\mathbf{0}\Bigr)
  \]
  \emph{(Only if anomaly pixels exist.)}
}
\State \quad \parbox[t]{\dimexpr\linewidth-\algorithmicindent}{
  \textbf{Anomaly recon loss}:
  \[
     \textit{rec\_anom}=\texttt{MSE}\bigl(\textit{recon}\times\textit{mask},\,\textit{img}\times\textit{mask}\bigr)
  \]
}
\State \quad \parbox[t]{\dimexpr\linewidth-\algorithmicindent}{
  \textbf{Total Variation}:
  \[
    \textit{tv\_loss}=\texttt{TV\_Loss}(\textit{recon})
  \]
  \emph{(If \textit{lambda\_tv} $>$ 0.)}
}
\State \quad \parbox[t]{\dimexpr\linewidth-\algorithmicindent}{
  \textbf{Color prior loss}:
  \begin{itemize}
    \item If \texttt{stageA} is provided, measure $\texttt{MSE}(\textit{recon}\times\textit{mask},\,\textit{stageA}\times\textit{mask})$.
    \item Else, compare to a constant color $\mathbf{c}_{\mathrm{ref}}$.
  \end{itemize}
}
\State \quad \parbox[t]{\dimexpr\linewidth-\algorithmicindent}{
  \textbf{Perceptual loss}:
  \[
    \textit{perc\_loss}=\|\psi(\textit{recon\_anom}) - \psi(\textit{img\_anom})\|^2;
  \]
}
\State \quad \parbox[t]{\dimexpr\linewidth-\algorithmicindent}{
  \textbf{Combine all sublosses:}
}
\State \quad \parbox[t]{\dimexpr\linewidth-\algorithmicindent}{
  \textbf{Backward + update}:
}
\State \quad \parbox[t]{\dimexpr\linewidth-\algorithmicindent}{
  Accumulate each subloss in running totals.
}
\State \parbox[t]{\dimexpr\linewidth-\algorithmicindent}{
\textbf{Compute average losses} $\{\textit{avg\_rec},\,\textit{avg\_pde},\dots\}$ after the loop, and return or log these metrics.
}
\end{algorithmic}
\end{algorithm}

In essence, \textbf{npcF} enforces a PDE-inspired correction (\textbf{Allen--Cahn}) to prevent anomaly regions from degenerating to trivial solutions (like pure black or oversmoothed patches). By simultaneously applying color and perceptual losses, we ensure that the synthesized anomalies maintain plausible tones and structures, thus forming a robust \emph{coarse refinement stage} before more detailed enhancements in subsequent pipelines.

\subsection{npcF++}
\label{sec:subsec_npcfpp}

Our \textit{npcF++} stage further refines anomalies by integrating \textbf{wavelet-domain PDE blocks} and \textbf{synergy cross-attention} across multiple resolution scales. Below, we split its training procedure into two parts: the core loop (\emph{Algorithm~\ref{alg:npcfpp_train}}) and an excerpt illustrating how \textsc{SynergyRefineNet.forward} combines normal/anomaly branches with boundary information (\emph{Algorithm~\ref{alg:npcfpp_forward}}). This design ensures boundary-sensitive, fine-grained anomaly generation.

\begin{algorithm}[H]
\caption{\textsc{npcF++} Training Flow}
\label{alg:npcfpp_train}
\textbf{Input:} \parbox[t]{\dimexpr\linewidth-\algorithmicindent}{
\textbf{SynergyRefineNet model} (dual-branch PDEWaveAttn + synergy),\\
Batches of \{\textit{b1}, \textit{orig}, \textit{mask}\}, \\
\textit{optimizer} (e.g.\ AdamW), wavelet penalty weight $\alpha_{\mathrm{wave}}$.
}
\textbf{Output:} \parbox[t]{\dimexpr\linewidth-\algorithmicindent}{
Updated model parameters (fine-grained refinement).
}
\begin{algorithmic}[1]

\State \parbox[t]{\dimexpr\linewidth-\algorithmicindent}{
  \textbf{function} \textsc{trainOneEpoch\_npcFpp}(\textit{model}, \textit{loader}, \textit{optimizer}):
}

\State \quad \parbox[t]{\dimexpr\linewidth-\algorithmicindent}{
  Initialize \(\textit{total\_loss}=0,\;\textit{count}=0.\)
}

\State \quad \parbox[t]{\dimexpr\linewidth-\algorithmicindent}{
  \textbf{for each mini-batch} \((\textit{b1},\,\textit{orig},\,\textit{mask})\) in \textit{loader}:
}
\State \qquad \parbox[t]{\dimexpr\linewidth-\algorithmicindent}{
  \(\textit{out}\gets\textit{model.forward}(\textit{orig},\,\textit{b1},\,\textit{mask})\).
}
\State \qquad \parbox[t]{\dimexpr\linewidth-\algorithmicindent}{
  \(\textit{reg}=\textsc{region\_loss}(\textit{out},\,\textit{orig},\,\textit{b1},\,\textit{mask})\).
}
\State \qquad \parbox[t]{\dimexpr\linewidth-\algorithmicindent}{
  \(\textit{wave}=\textsc{wave\_hf\_loss}(\textit{out},\,\textit{mask})\).
}
\State \qquad \parbox[t]{\dimexpr\linewidth-\algorithmicindent}{
  \(\textit{loss}\gets \textit{reg}\;+\;\alpha_{\mathrm{wave}}\,\times\,\textit{wave}\).
}
\State \qquad \parbox[t]{\dimexpr\linewidth-\algorithmicindent}{
  \(\textit{total\_loss} \gets \textit{total\_loss} + (\textit{loss}\times\textit{batch\_size});\\
  \;\textit{count}\gets \textit{count} + \textit{batch\_size}.\)
}

\State \quad \parbox[t]{\dimexpr\linewidth-\algorithmicindent}{
  \(\textbf{return}\;\textit{total\_loss}/\textit{count}\).
}
\end{algorithmic}
\end{algorithm}

\begin{algorithm}[H]
\caption{\textsc{SynergyRefineNet.forward}}
\label{alg:npcfpp_forward}
\textbf{Input:} \parbox[t]{\dimexpr\linewidth-\algorithmicindent}{
\textit{orig} (normal input), \textit{b1} (coarse anomaly), \textit{mask} (binary), \\
Multi-scale PDEWaveAttn \& synergy blocks.
}
\textbf{Output:} \parbox[t]{\dimexpr\linewidth-\algorithmicindent}{
\textit{out} (refined anomaly, shape $(B,3,H,W)$).
}
\begin{algorithmic}[1]

\State \parbox[t]{\dimexpr\linewidth-\algorithmicindent}{
  \textbf{Encode normal:}\quad
  \(\begin{aligned}[t]
    &eN1=\mathrm{EncN1}(\textit{orig});\quad pN1=\mathrm{PoolN1}(eN1);\\
    &eN2=\mathrm{EncN2}(pN1);\quad pN2=\mathrm{PoolN2}(eN2);\\
    &bN=\mathrm{BottN}(pN2).
  \end{aligned}\)
}

\State \parbox[t]{\dimexpr\linewidth-\algorithmicindent}{
  \textbf{Encode anomaly:}\quad
  \(\begin{aligned}[t]
    &eA1=\mathrm{EncA1}(\textit{b1});\quad pA1=\mathrm{PoolA1}(eA1);\\
    &eA2=\mathrm{EncA2}(pA1);\quad pA2=\mathrm{PoolA2}(eA2);\\
    &bA=\mathrm{BottA}(pA2).
  \end{aligned}\)
}

\State \parbox[t]{\dimexpr\linewidth-\algorithmicindent}{
  \textbf{Decode (scale 1) + synergy:}\quad
  \(\begin{aligned}[t]
    &upN1=\mathrm{upN1}(bN);\\
    &upA1=\mathrm{upA1}(bA);\\
    &dN1=\mathrm{decN1}(upN1,\dots,eN2);\\
    &dA1=\mathrm{decA1}(upA1,\dots,eA2);\\
    &synergy1=\mathrm{synergy1}(dN1,\,dA1,\,\textit{mask}).
  \end{aligned}\)
}

\State \parbox[t]{\dimexpr\linewidth-\algorithmicindent}{
  \textbf{Decode (scale 2) + synergy:}\quad
  \(\begin{aligned}[t]
    &upN2=\mathrm{upN2}(synergy1);\\
    &upA2=\mathrm{upA2}(synergy1);\\
    &dN2=\mathrm{decN2}(upN2,\dots,eN1);\\
    &dA2=\mathrm{decA2}(upA2,\dots,eA1);\\
    &synergy2=\mathrm{synergy2}(dN2,\,dA2,\,\textit{mask}).
  \end{aligned}\)
}

\State \parbox[t]{\dimexpr\linewidth-\algorithmicindent}{
  \textbf{Final:}
  \[
    \textit{out}=\texttt{sigmoid}\bigl(\mathrm{final}(synergy2)\bigr).
  \]
  \textbf{return} \(\textit{out}\).
}
\end{algorithmic}
\end{algorithm}

In summary, \textbf{npcF++} extends beyond \textit{npcF} by adding wavelet-based PDE blocks for multi-scale refinement and synergy cross-attention to fuse normal/anomaly features. This results in more coherent boundaries and a stronger perceptual quality in synthesized anomalies.

\subsection{BiSQAD}
\label{sec:subsec_bisqad}

Our \emph{BiSQAD} framework combines a \textbf{Synthesis Quality Estimator (SQE)} with a \textbf{two-loop} optimization strategy to dynamically reweight synthetic anomalies based on their relevance to the end detection task. Below, we present two key pseudocode snippets: (1)~the \textit{inner loop} that updates the main anomaly detection model and collects per-sample losses, and (2)~the \textit{outer loop} that refines either data weights or SQE parameters via second-order gradient steps.  

\begin{algorithm}[H]
\caption{\textsc{trainOneEpoch\_BiSQAD} (Inner Loop)}
\label{alg:bisqad_inner}
\textbf{Input:} \parbox[t]{\dimexpr\linewidth-\algorithmicindent}{
\textbf{full\_model} (anomaly detector), \\
\textbf{train\_loader} w/ \(\{\,\textit{image}, \textit{label}, \textit{img\_sqe}, \textit{index}\}\), \\
\textbf{optimizer}, \\
\textit{data\_weights}, \(\textit{sqe\_model}\) (optional), \(\lambda_{\mathrm{sqe}}, \lambda_{\mathrm{bi}}\).
}
\textbf{Output:} \parbox[t]{\dimexpr\linewidth-\algorithmicindent}{
Updated \textbf{full\_model}, and arrays of \{\textit{index}, \\
\textit{loss\_value}, \textit{img\_sqe}\} for subsequent outer loop \\
training.
}
\begin{algorithmic}[1]

\State \parbox[t]{\dimexpr\linewidth-\algorithmicindent}{
  \textbf{function} \textsc{trainOneEpoch\_BiSQAD}(
       \textit{config}, \textit{train\_loader}, \textit{full\_model}, 
       \textit{optimizer}, \textit{data\_weights}, \textit{sqe\_model}):
}

\State \quad \parbox[t]{\dimexpr\linewidth-\algorithmicindent}{
  Initialize accumulators: \(\{\textit{epoch\_indices},\,\textit{epoch\_losses},\,\textit{epoch\_imgs\_sqe}\}\gets\emptyset\).
}

\State \quad \parbox[t]{\dimexpr\linewidth-\algorithmicindent}{
  \textbf{for each mini-batch} \((\textit{batch})\) in \textit{train\_loader}:
}
\State \qquad \parbox[t]{\dimexpr\linewidth-\algorithmicindent}{
  $\textit{outputs}\gets \textit{full\_model.forward}(\textit{batch},\,train=\text{True})$
}

\State \qquad \parbox[t]{\dimexpr\linewidth-\algorithmicindent}{
  \textbf{Compute loss per sample:}\quad
  \(\begin{aligned}[t]
    \textit{sum\_loss\_b} \;\gets\;&\sum_{\textit{criterion}\in \mathcal{C}}\bigl(
    \textit{criterion.weight}\;\times \\
    &\textit{criterion}(\textit{outputs},\,\textit{per\_sample=True})
    \bigr).
  \end{aligned}\)
}
\State \qquad \parbox[t]{\dimexpr\linewidth-\algorithmicindent}{
  \textbf{Get SQE or data weights}:
  \begin{itemize}
    \item If \(\textit{sqe\_model}\) is not None, \(\textit{sqe\_score}\gets \textit{sqe\_model.forward}(\textit{batch.img\_sqe})\)
    \item Else, \(\textit{sqe\_score}\gets \mathbf{0}\)
    \item \(\textit{meta\_w}\gets\textit{data\_weights}[\textit{batch.index}]\)
    \item \(\textit{final\_w}\gets\lambda_{\mathrm{sqe}}\cdot \textit{sqe\_score} + \lambda_{\mathrm{bi}}\cdot \textit{meta\_w}\)
    \item \emph{If \(\lambda_{\mathrm{sqe}}{=}0\), fallback to uniform weighting.}
  \end{itemize}
}
\State \qquad \parbox[t]{\dimexpr\linewidth-\algorithmicindent}{
  \textbf{Weighted total:}\quad
  \(\begin{aligned}[t]
    &\textit{weighted\_loss\_b} = \\ 
    &\textit{sum\_loss\_b}\times \textit{final\_w},\\
    &\textit{total\_loss}\gets \textit{weighted\_loss\_b.mean()}.
  \end{aligned}\)
}

\State \qquad \parbox[t]{\dimexpr\linewidth-\algorithmicindent}{
  \textbf{Backward + optimize:}
}

\State \qquad \parbox[t]{\dimexpr\linewidth-\algorithmicindent}{
  \textbf{Accumulate for outer loop:}\quad
  \(\begin{aligned}[t]
    &\textit{epoch\_indices}\mathrel{+}= \textit{batch.index},\\[2pt]
    &\textit{epoch\_losses}\mathrel{+}= \textit{sum\_loss\_b.detach()},\\[2pt]
    &\textit{epoch\_imgs\_sqe}\mathrel{+}= \textit{batch.img\_sqe}.
  \end{aligned}\)
}

\State \quad \parbox[t]{\dimexpr\linewidth-\algorithmicindent}{
  \textbf{return} \(\{\textit{epoch\_indices},\,\textit{epoch\_losses},\,\textit{epoch\_imgs\_sqe}\}\) for outer loop usage.
}

\end{algorithmic}
\end{algorithm}

\begin{algorithm}[H]
\caption{\textsc{OuterLoop\_SQE} (Second-Order Update)}
\label{alg:bisqad_outer}
\textbf{Input:} \parbox[t]{\dimexpr\linewidth-\algorithmicindent}{
\textbf{full\_model}, \textbf{data\_weights}, \textbf{sqe\_model}, \\
\textbf{train\_batch} (a new sample), \textbf{val\_loader}, \\
\textbf{optimizer\_w} (for \textit{data\_weights}), \\
\textbf{config} (\eg, \(\textit{meta\_inner\_lr},\,\textit{auc\_alpha}\)).
}
\textbf{Output:} \parbox[t]{\dimexpr\linewidth-\algorithmicindent}{
Updated \(\textit{data\_weights}\) and \(\textit{sqe\_model}\) reflecting second-order optimization.
}

\begin{algorithmic}[1]
\State \parbox[t]{\dimexpr\linewidth-\algorithmicindent}{
   \textbf{function} \textsc{outerLoopSQE}(\textit{config}, \textit{full\_model}, \textit{data\_weights}, \textit{optimizer\_w}, \textit{train\_batch}, \textit{val\_loader}):
}
\State \quad \parbox[t]{\dimexpr\linewidth-\algorithmicindent}{
  \textbf{Create temp copy}: $\textit{temp\_model}\gets \texttt{copy.deepcopy}(\textit{full\_model})$
}
\State \quad \parbox[t]{\dimexpr\linewidth-\algorithmicindent}{
  \textbf{Inner Step}:
  \begin{enumerate}
  \item Retrieve $\textit{final\_w}\gets\textit{data\_weights}[\textit{train\_batch.index}]$
  \item $\textit{inner\_loss}\gets \textit{temp\_model}(\textit{train\_batch})\times \textit{final\_w}$
  \item Update $\textit{temp\_model}$ w.r.t.\ $\textit{inner\_loss}$ (using $\textit{meta\_inner\_lr}$).
  \end{enumerate}
}
\State \quad \parbox[t]{\dimexpr\linewidth-\algorithmicindent}{
  \textbf{Val AUC Loss:}
  \begin{itemize}
    \item Evaluate \(\textit{temp\_model}\) on \(\texttt{val\_loader}\), gather \(\textit{val\_scores}\), \(\textit{val\_labels}\).
    \item Compute AUC loss:\\[2pt]
    \(\textit{auc\_loss}\gets\mathrm{differentiable\_auc\_loss}(\textit{val\_scores},\,\\
    \textit{val\_labels},\,\alpha=\textit{auc\_alpha})\).
  \end{itemize}
}

\State \quad \parbox[t]{\dimexpr\linewidth-\algorithmicindent}{
  \emph{(Updates both $\textit{data\_weights}$ \& final layer of $\textit{sqe\_model}$ if attached.)}
}
\State \quad \parbox[t]{\dimexpr\linewidth-\algorithmicindent}{
  \textbf{Discard} \(\textit{temp\_model}\), retaining only updated \(\textit{data\_weights}, \textit{sqe\_model}\).
}
\end{algorithmic}
\end{algorithm}

\textbf{Explanation.} In \cref{alg:bisqad_inner}, we collect per-sample losses in the \emph{inner loop}, weighting them with a combination of \(\lambda_{\mathrm{sqe}}\cdot\textit{sqe\_score}\) and \(\lambda_{\mathrm{bi}}\cdot\textit{data\_weights}\). Then, \cref{alg:bisqad_outer} runs a second-order update (inspired by \emph{MAML}) that \emph{locally} adapts a \(\textit{temp\_model}\), measures validation-based \(\textit{auc\_loss}\), and backpropagates to refine \(\textit{data\_weights}\). This procedure selectively amplifies beneficial anomalies while reducing the impact of low-fidelity samples, hence improving overall detection performance.

\section{Additional Details of the \textit{MaPhC2F} Dataset}
\label{sec:maphc2f_supp}

In the main paper, we introduce the \textbf{\textit{MaPhC2F Dataset}}, a large-scale synthetic anomaly dataset spanning multiple object categories across \emph{MVTec~AD}, \emph{VisA}, and \emph{BTAD}. Here, we provide further details, including a per-category breakdown of the number of images generated for each mechanism of synthetic defect, and illustrative examples showing how \textit{MaPhC2F} enriches existing industrial datasets with diverse anomalies.

\subsection{Data Quantities per Defect Mechanism}
\label{sec:maphc2f_table}

Table~\ref{tab:maphc2f_mvtec}, ~\ref{tab:maphc2f_visa},  ~\ref{tab:maphc2f_btad} enumerates the total image counts for each defect category and each object class. We note that our pipeline systematically applies three physically inspired processes (Fracture-Line, Pitting-Loss, TPS-based Plastic-Warp) to generate varied and realistic anomaly masks. After refinement (\textit{npcF} and \textit{npcF++}), we accumulate the resulting synthesized images under the \textbf{\textit{MaPhC2F Dataset}}. 

\begin{table*}
\centering
\resizebox{\linewidth}{!}{
\begin{tabular}{l|ccccccccccccccc}
\toprule
\textbf{\textit{Defect Mechanism}} 
& \textbf{\textit{Bottle}} & \textbf{\textit{Cable}} & \textbf{\textit{Capsule}} & \textbf{\textit{Carpet}} & \textbf{\textit{Grid}} & \textbf{\textit{Hazelnut}} & \textbf{\textit{Leather}} & \textbf{\textit{Metal\_Nut}} & \textbf{\textit{Pill}} & \textbf{\textit{Screw}} & \textbf{\textit{Tile}} & \textbf{\textit{Toothbrush}} & \textbf{\textit{Transistor}} & \textbf{\textit{Wood}} & \textbf{\textit{Zipper}} \\
\midrule
\textit{Fracture-Line} 
& 627 & 672 & 657 & 840 & 792 & 1173 & 735 & 660 & 801 & 960 & 690 & 180 & 639 & 741 & 720 \\
\textit{Pitting-Loss}
& 627 & 672 & 657 & 840 & 792 & 1173 & 735 & 660 & 801 & 960 & 690 & 180 & 639 & 741 & 720 \\
\textit{Plastic-Warp}
& 627 & 672 & 657 & 840 & 792 & 1173 & 735 & 660 & 801 & 960 & 690 & 180 & 639 & 741 & 720 \\
\midrule
\textbf{Total}
& 1881& 2016& 1971 & 2520 & 2376 & 3519 & 2205 & 1980 & 2403 & 2880 & 2070 & 540 & 1917 & 2223 & 2160 \\
\bottomrule
\end{tabular}
}
\caption{Number of \textit{MaPhC2F} images on \textbf{\textit{MVTec~AD}} (15 categories). Rows list defect mechanisms and total counts; columns list the individual categories.}
\label{tab:maphc2f_mvtec}
\end{table*}

\begin{table*}
\centering
\resizebox{\linewidth}{!}{
\begin{tabular}{l|cccccccccccc}
\toprule
\textbf{\textit{Defect Type}} 
& \textbf{\textit{Candle}} & \textbf{\textit{Capsules}} & \textbf{\textit{Cashew}} & \textbf{\textit{Chewinggum}} & \textbf{\textit{Fryum}} 
& \textbf{\textit{Macaroni1}} & \textbf{\textit{Macaroni2}} & \textbf{\textit{PCB1}} & \textbf{\textit{PCB2}} & \textbf{\textit{PCB3}} & \textbf{\textit{PCB4}} & \textbf{\textit{Pipe\_Fryum}}\\
\midrule
\textit{Fracture-Line}
& 2700 & 1626 & 1350 & 1359 & 1350 & 2700 & 2700 & 2712 & 2703 & 2715 & 2712 & 1350 \\
\textit{Pitting-Loss}
& 2700 & 1626 & 1350 & 1359 & 1350 & 2700 & 2700 & 2712 & 2703 & 2715 & 2712 & 1350 \\
\textit{Plastic-Warp}
& 2700 & 1626 & 1350 & 1359 & 1350 & 2700 & 2700 & 2712 & 2703 & 2715 & 2712 & 1350 \\
\midrule
\textbf{\textit{Total}}
& 8100& 4878& 4050& 4077& 4050& 8100& 8100& 8136& 8109& 8145& 8136& 4050\\
\bottomrule
\end{tabular}
}
\caption{Number of \textit{MaPhC2F} images on \textbf{\textit{VisA}} (12 categories). Rows list defect mechanisms and total counts; columns list the individual categories.}
\label{tab:maphc2f_visa}
\end{table*}

\begin{table*}
\centering
\begin{tabular}{l|ccc}
\toprule
\textbf{Defect Type} & \textbf{01} & \textbf{02} & \textbf{03} \\
\midrule
Pitting-Loss & 400 & 399 & 1000 \\
Plastic-Warp       & 400 & 399 & 998  \\
Fracture-Line     & 400 & 399 & 1000 \\
\midrule
\textbf{Total} & 1200 & 1197 & 2998 \\
\bottomrule
\end{tabular}
\caption{Number of \textit{MaPhC2F} images on \textbf{BTAD} (3 categories). Rows list defect types and total counts; columns list the individual categories.}
\label{tab:maphc2f_btad}
\end{table*}

\subsection{Visualization of Synthetic Anomalies}
\label{sec:maphc2f_visual}

\begin{figure*}
    \centering
    \includegraphics[height=0.99\textheight]{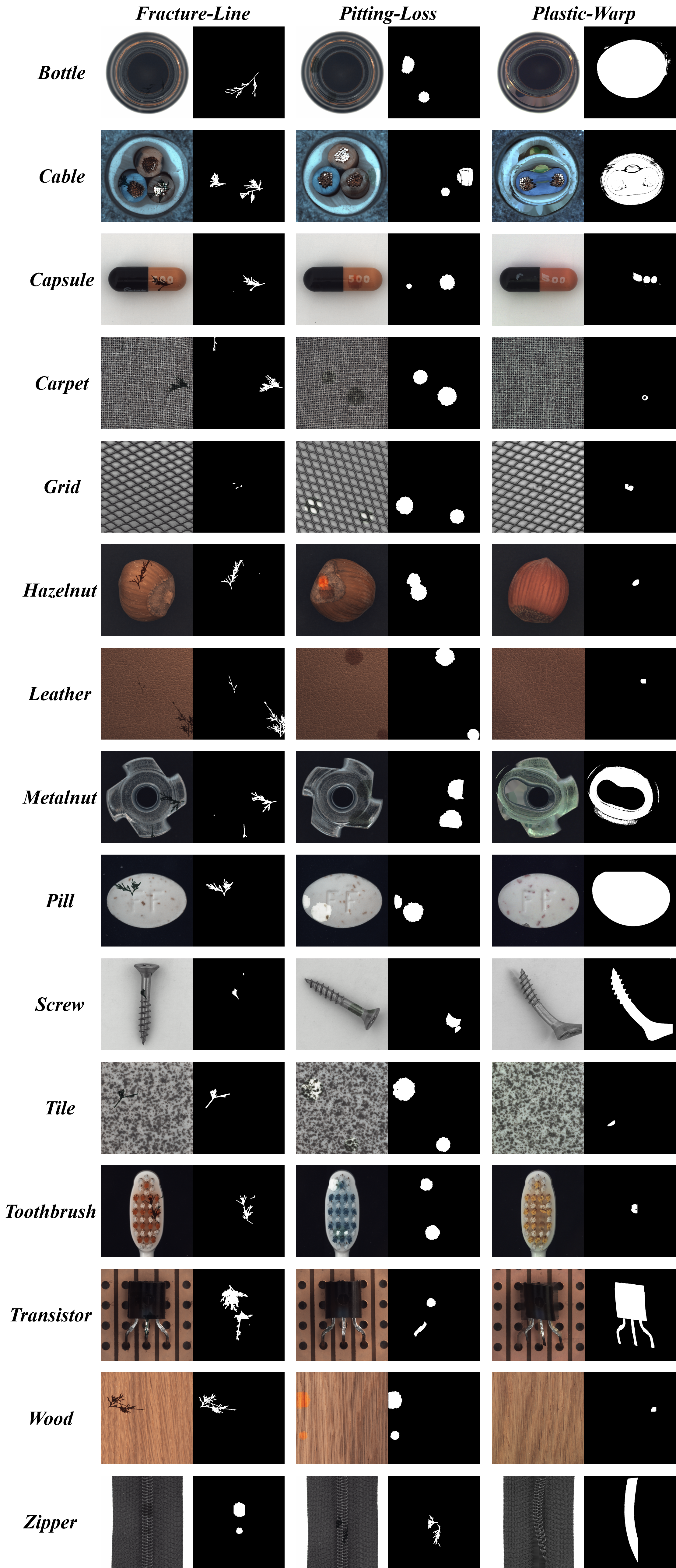}
    \caption{Examples of \textbf{\textit{MaPhC2F}} anomalies on \emph{MVTec~AD} categories. Each row corresponds to one item mechanism, and columns display Fracture-Line, Pitting-Loss, and Plastic-Warp, respectively.}
    \label{fig:mvtec_vis}
\end{figure*}

\begin{figure*}
    \centering
    \includegraphics[height=0.99\textheight]{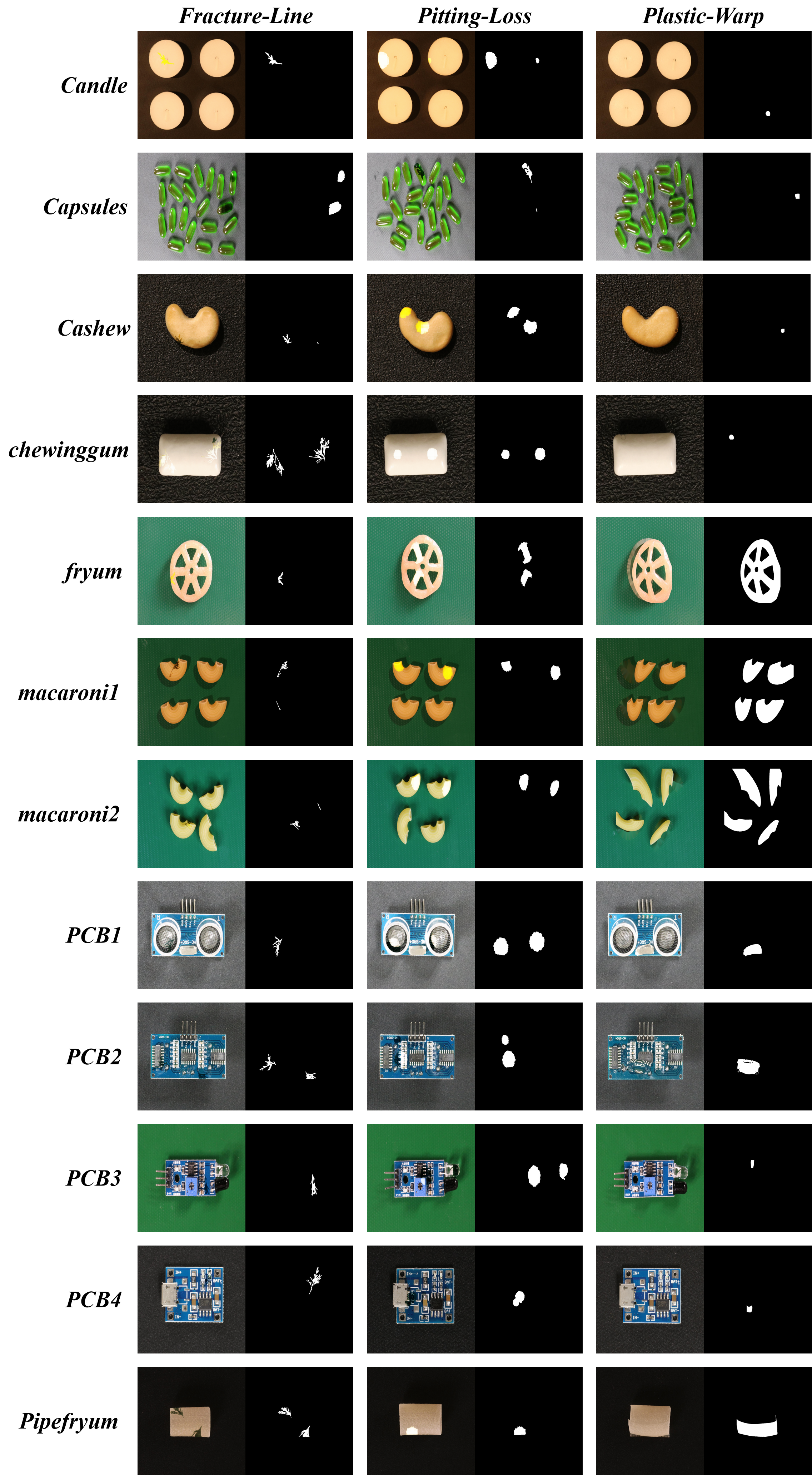}
    \caption{Anomalies synthesized for \emph{VisA} dataset classes, visualized in the same row-by-column arrangement as Figure~\ref{fig:mvtec_vis}.}
    \label{fig:visa_vis}
\end{figure*}

\begin{figure*}
    \centering
    \includegraphics[width=0.95\linewidth]{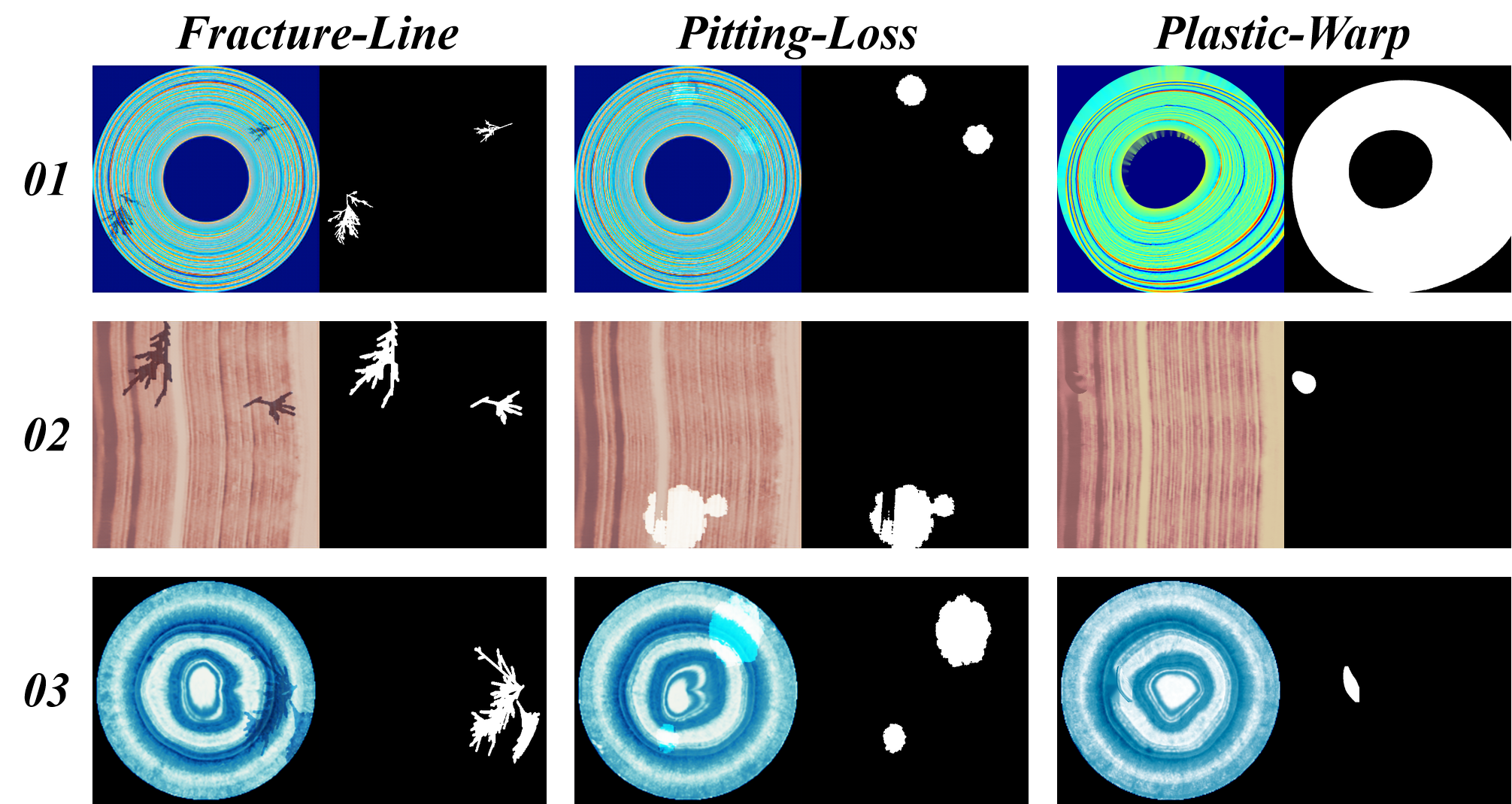}
    \caption{Additional \textbf{\textit{MaPhC2F}} samples on \emph{BTAD}, illustrating the consistency of our method across varied industrial items.}
    \label{fig:btad_vis}
\end{figure*}

Fig.~\ref{fig:mvtec_vis}, Fig.~\ref{fig:visa_vis}, and Fig.~\ref{fig:btad_vis} depict representative samples of \textbf{\textit{MaPhC2F}} images for \emph{MVTec~AD}, \emph{VisA}, and \emph{BTAD} categories, respectively. Each figure arranges one category per row and shows three columns corresponding to the three synthetic anomaly mechanisms:

\begin{itemize}
    \item \textbf{Fracture-Line:} Masks grown via distance transforms and Perlin noise, emulating fracture lines.
    \item \textbf{Pitting-Loss:} Chunky, “rust-like” patches with edge disruption.
    \item \textbf{Plastic-Warp:} TPS-based local warping or dent-like Plastic-Warp.
\end{itemize}

Overall, the \textbf{\textit{MaPhC2F Dataset}} contains over \textit{115,987} synthetic images across 30 categories, serving as a valuable resource for anomaly detection. Its physically guided design yields anomalies whose shape and texture exhibit realistic correlations to real defects, while the integrated coarse-to-fine refinement process preserves rich details for training more robust detection models.

\section{Qualitative Results}
\label{sec:qualitative_results}

\begin{figure*}
    \centering
    \includegraphics[width=\linewidth]{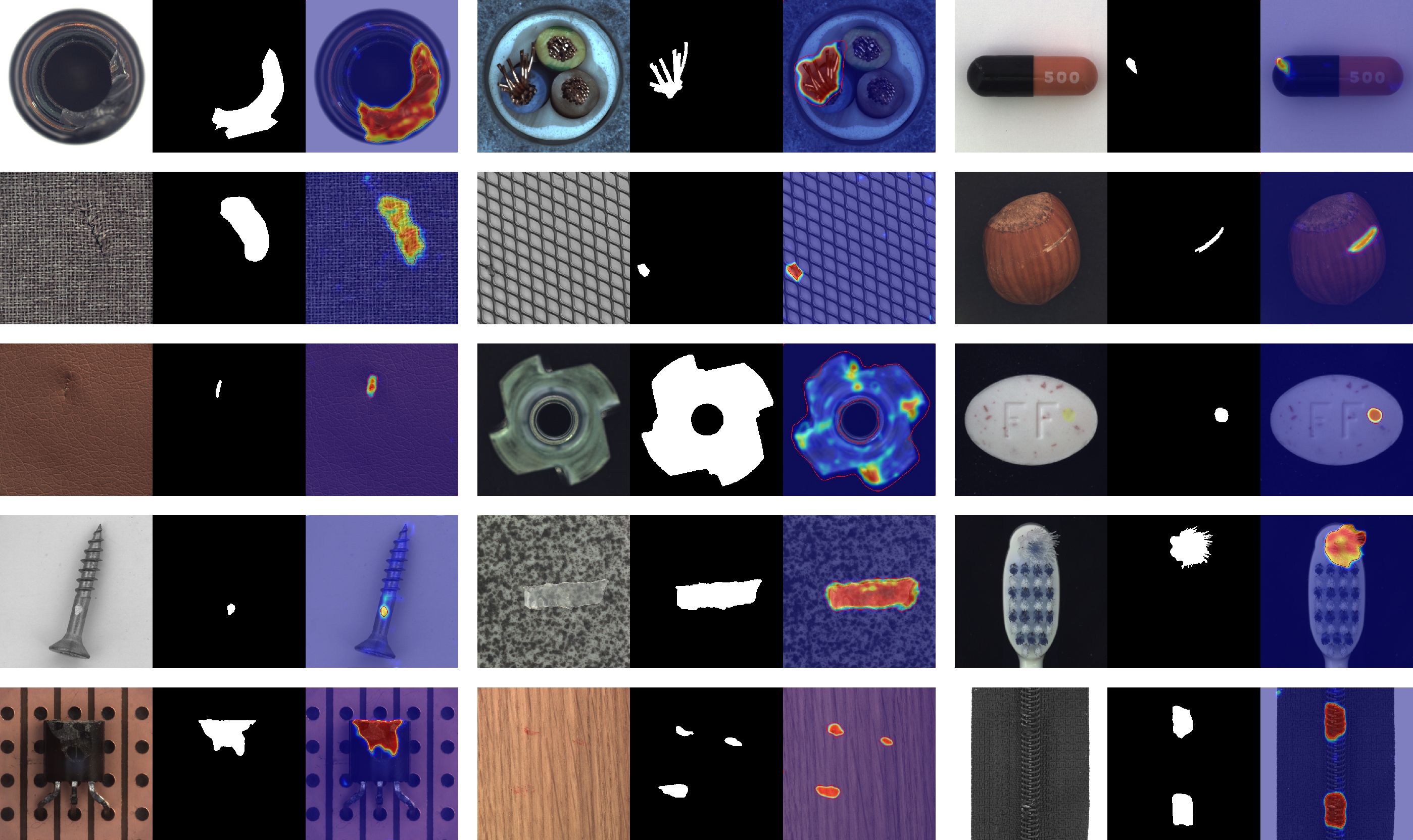}
    \caption{Visual examples of anomaly detection on several \textit{MVTec~AD} categories.
    Each group shows: (1) original input, (2) ground-truth mask, (3) overlay of detection results.}
    \label{fig:mvtec_res}
\end{figure*}

\begin{figure*}
    \centering
    \includegraphics[width=\linewidth]{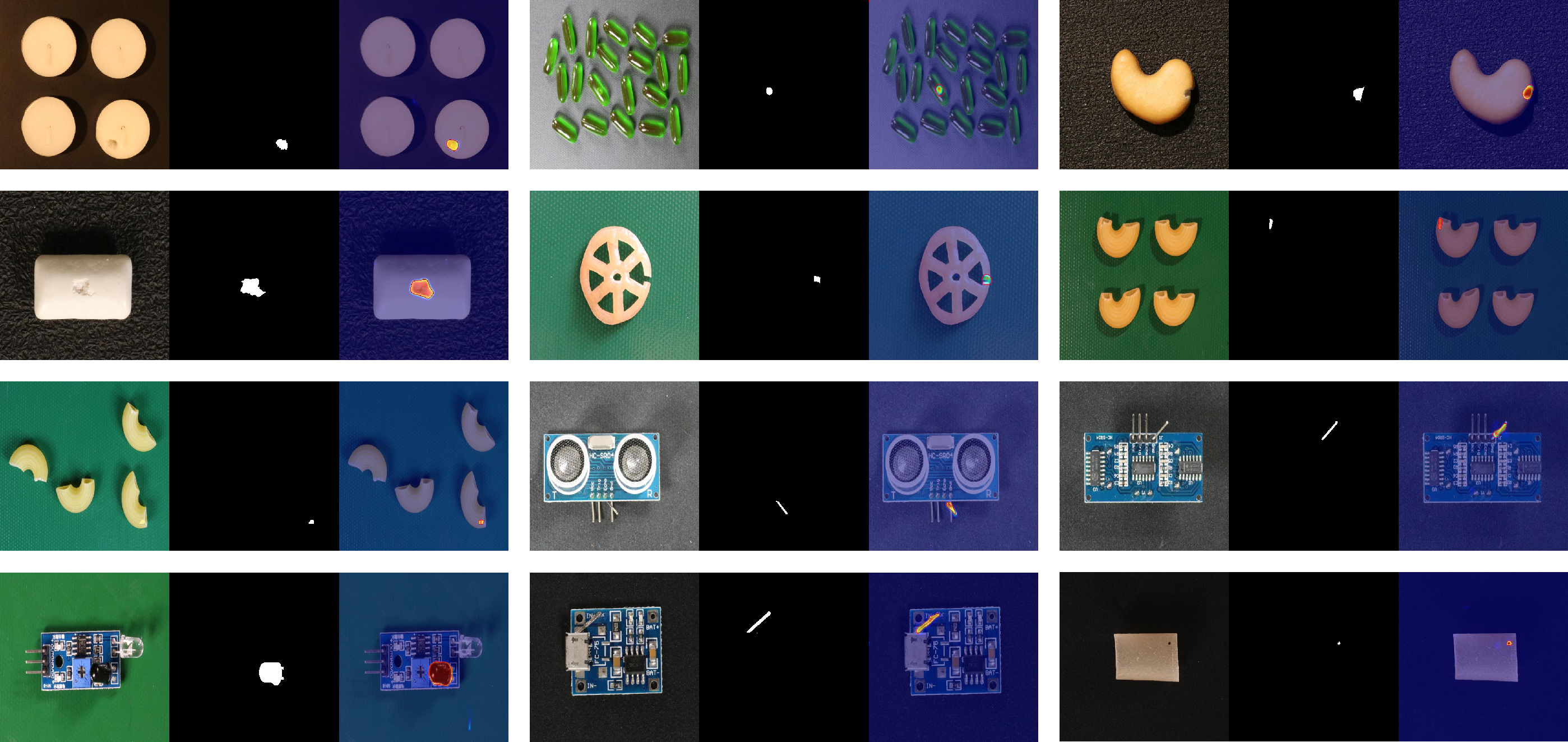}
    \caption{Visualization of detection outcomes on various \textit{VisA} items. Our method adapts well to diverse categories.}
    \label{fig:visa_res}
\end{figure*}

\begin{figure*}
    \centering
    \includegraphics[width=\linewidth]{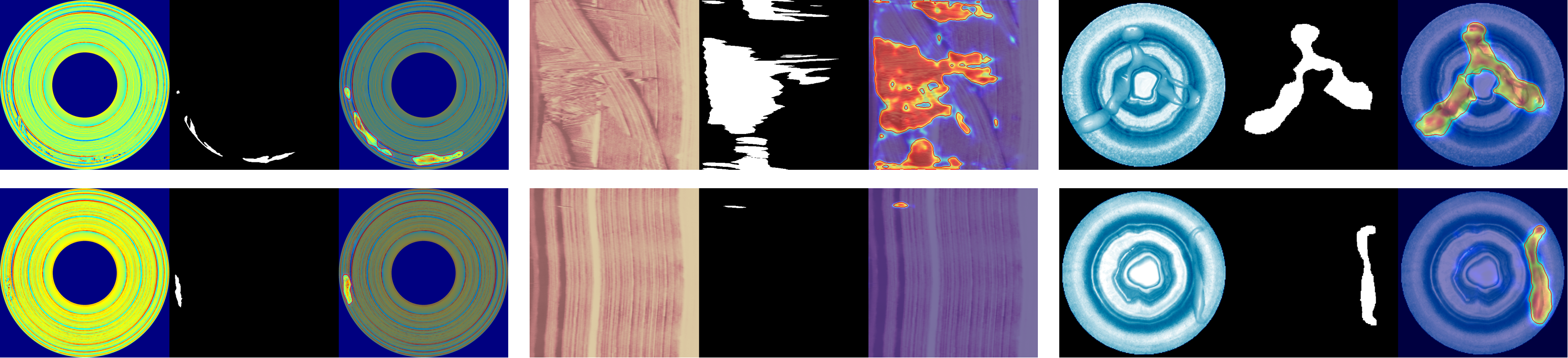}
    \caption{Examples of anomaly detection on \textit{BTAD} data.}
    \label{fig:btad_res}
\end{figure*}

In addition to the quantitative improvements demonstrated by our approach, we present a set of visual comparisons on \textit{MVTec~AD}, \textit{VisA}, and \textit{BTAD} to illustrate how \textbf{MaPhC2F + BiSQAD} can detect and localize anomalies more effectively, as shown in Fig.~\ref{fig:mvtec_res}, ~\ref{fig:visa_res}, ~\ref{fig:btad_res}.

Across \textit{MVTec~AD}, \textit{VisA}, and \textit{BTAD}, these qualitative examples reinforce the quantitative findings presented earlier, illustrating that \textbf{MaPhC2F + BiSQAD} achieves precise, fine-grained anomaly localization even under challenging scenarios.

\section{Broader Impact and Ethics}
\label{sec:broader}

\paragraph{Positive impact.}
Open-sourcing the \textit{MaPhC2F} generators together with a 115\,987-image dataset lets small manufacturers and research labs train anomaly detectors without collecting thousands of real defects, thereby reducing scrap, energy use, and operator exposure to hazardous production lines.  The entire pipeline runs at 42 FPS, so it can be deployed on edge devices for real-time quality assurance in resource-constrained settings.

\paragraph{Potential misuse and mitigations.}
\begin{itemize}
    \item \emph{Fabricated evidence.}  Synthetic defects could be used to forge warranty claims or falsify inspection logs. 
    \item \emph{Disinformation imagery.}  The generator might be repurposed to create alarming “failure” photos on social media. 
\end{itemize}

\paragraph{Mitigation:} the code is released under CC-BY-NC-4.0, prohibiting commercial use, and the repository includes an explicit policy restricting use to academic/industrial inspection research.  
We believe the benefits—safer, cheaper, and more sustainable visual inspection—outweigh these manageable risks.

\end{document}